\documentclass[conference]{IEEEtran}

\IEEEoverridecommandlockouts
\usepackage{cite}
\usepackage{amsmath,amssymb,amsfonts}
\usepackage{algorithmic}
\usepackage{graphicx}
\usepackage{textcomp}
\usepackage{xcolor}

\def\BibTeX{{\rm B\kern-.05em{\sc i\kern-.025em b}\kern-.08em
    T\kern-.1667em\lower.7ex\hbox{E}\kern-.125emX}}
\usepackage{tikz}
\usepackage{url}
\usepackage{hyperref} 
\hypersetup{hidelinks,
    colorlinks=true,
    allcolors=black,
    pdfstartview=Fit,
    breaklinks=true
}
\usepackage{siunitx}
\usepackage{threeparttable}
\usepackage{multirow}
\usepackage{makecell}
\usepackage{booktabs}
\usepackage{float}
\usepackage{gensymb}
\usepackage[linesnumbered,ruled,vlined]{algorithm2e}
\usepackage{graphicx}
\usepackage{subcaption}

\newcommand{\revision}[1]{\textcolor{black}{#1}} 

\SetCommentSty{emph} 
\newcommand{\mycomment}[1]{\textcolor{green!50!black}{#1}}

\begin{document}
\title{
Improving Robustness to Out-of-Distribution States in Imitation Learning via Deep Koopman-Boosted Diffusion Policy
}

\author{
    % \vskip 1em
    Dianye Huang, \emph{Graduate Student Member, IEEE},
    Nassir Navab, \emph{Fellow, IEEE},
    Zhongliang Jiang \emph{Member, IEEE}
    \thanks{
		% Manuscript received Month xx, 2xxx; revised Month xx, xxxx; accepted Month x, xxxx.
		% This work was supported in part by the xxx Department of xxx under Grant  (sponsor and financial support acknowledgment goes here).
        \textit{Corresponding author: Zhongliang Jiang}
        
        Dianye Huang, Nassir Navab, and Zhongliang Jiang are with the Chair for Computer Aided Medical Procedures and Augmented Reality (CAMP), Technical University of Munich (TUM), Boltzmannstr. 3, 85748 Garching bei München, Germany; Munich Center for Machine Learning (MCML). (e-mail: dianye.huang@tum.de, nassir.navab@tum.de, zl.jiang@tum.de). Zhongliang Jiang is also with the Medical Intelligence and Robotic Cognition Lab (MIRoC), The University of Hong Kong (HKU), Hong Kong, China (email: zljiang@hku.hk).
    }
}

\maketitle

\begin{abstract}
Integrating generative models with action chunking has shown significant promise in imitation learning for robotic manipulation. However, the existing diffusion-based paradigm often struggles to capture strong temporal dependencies across multiple steps, particularly when incorporating proprioceptive input. This limitation can lead to task failures, where the policy overfits to proprioceptive cues at the expense of capturing the visually derived features of the task. To overcome this challenge, we propose the Deep Koopman-boosted Dual-branch Diffusion Policy (D3P) algorithm. D3P introduces a dual-branch architecture to decouple the roles of different sensory modality combinations. The visual branch encodes the visual observations to indicate task progression, while the fused branch integrates both visual and proprioceptive inputs for precise manipulation. Within this architecture, when the robot fails to accomplish intermediate goals, such as grasping a drawer handle, the policy can dynamically switch to execute action chunks generated by the visual branch, allowing recovery to previously observed states and facilitating retrial of the task. To further enhance visual representation learning, we incorporate a Deep Koopman Operator module that captures structured temporal dynamics from visual inputs. During inference, we use the test-time loss of the generative model as a confidence signal to guide the aggregation of the temporally overlapping predicted action chunks, thereby enhancing the reliability of policy execution. In simulation experiments across six RLBench tabletop tasks, D3P outperforms the state-of-the-art diffusion policy by an average of 14.6\%. On three real-world robotic manipulation tasks, it achieves a 15.0\% improvement. Code: \url{https://github.com/dianyeHuang/D3P}. 
\end{abstract}


%%%%%%%%%%%%%%%%%% main content
% https://www.tablesgenerator.com/#  % create table

\section{Introduction}
\label{sec:intro}
% background of imitation learning and states the philosophy behind
\IEEEPARstart{I}{mitation} learning (IL), a simple yet effective and scalable learning paradigm for sequential decision-making tasks, has demonstrated remarkable success in diverse applications such as gaming AI, autonomous driving, and robotic manipulation
% ~\cite{ajak2024comparison, xie2024decomposing}. 
~\cite{park2021object, ajak2024comparison, xie2024decomposing}. 
Fundamentally, IL follows a supervised learning framework where the agent learns complex behaviors by mapping expert demonstration states to corresponding actions without enforcing direct interaction with the environment. 
% introduce action chunkings 
Despite its effectiveness, a conventional IL policy with a single action as output may struggle to capture long-term temporal dependencies between actions~\cite{chi2023diffusion}, leading to issues such as getting stuck in certain states during deployment, for instance, due to \textit{idle states} in the demonstrations of a liquid pouring task. To address this challenge and reduce the \revision{covariate} shift due to the compounding errors, a recent paradigm combining generative model and action chunking~\cite{lai2022action} has gained increasing attention~\cite{chi2023diffusion, zhao2023learning}. In such a paradigm, the generative model~\cite{vaswani2017attention, ho2020denoising} accommodates the action chunkings (ACs), which group multiple actions together, from expert demonstrations that vary in style.
% , and ACs, which group multiple actions together, can be viewed as motion primitives that constitute the demonstrated action trajectories. 
During deployment, the learned policy encodes the observation from the environment as contextual information, which the generative model then utilizes to generate AC containing continuous multi-step actions for execution. 

\begin{figure}[t!]
\centering
\includegraphics[width=0.99\columnwidth]{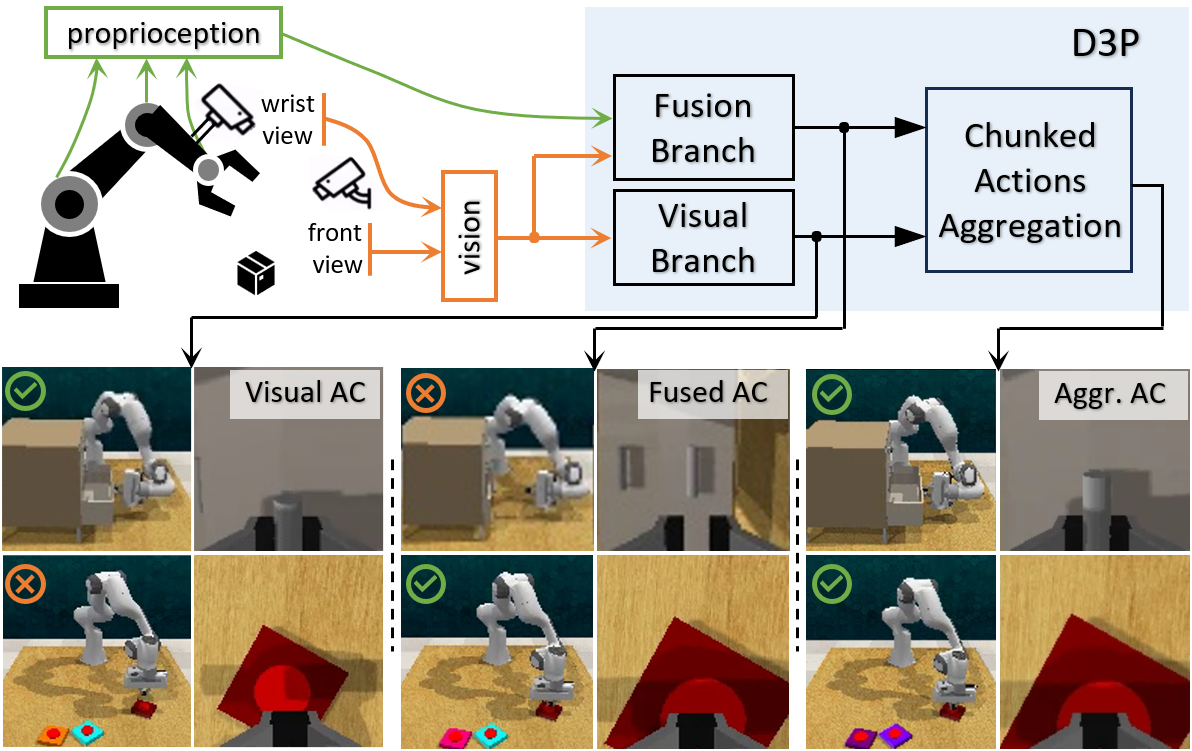}
\caption{An illustration of the proposed D3P algorithm, featuring a dual-branch architecture that generates two ACs at each inference step, and an aggregation module that synthesizes the final action sequence based on the \textit{test-time loss} of the generative model.}
\label{fig:teaser}
\end{figure}

% discussion on the causality problem
\par
However, the generative approach equipped with AC inherits several limitations of IL, most notably the issues of causal confusion and copycat behaviors~\cite{de2019causal, wen2020fighting, chuang2022resolving}. These issues arise when the policy fails to distinguish between merely correlational patterns and true causal relationships in expert demonstrations. As a result, it may blindly imitate observed action sequences without understanding their underlying purpose. 
% motivation and summarize the contributions
As illustrated in Fig. \ref{fig:illustration}, we trained the Diffusion Policy~\cite{chi2023diffusion} using both visual and proprioceptive data, as well as using visual data alone for comparison. For the Open Drawer task, we observed that the policy trained with both visual and proprioceptive data tends to over-rely on proprioception. Consequently, it struggles to recover from out-of-distribution joint states, often getting stuck in oscillatory behavior when the robotic arm fails to grasp the drawer handle. This behavior is consistent with the findings of shortcut learning reported by Geirhos~\emph{et al.} in~\cite{geirhos2020shortcut} that the DNNs tend to favor the simpler, shortcut solution over investing effort to uncover the intended, more complex representations. In contrast, the policy trained with visual input alone exhibits unexpected failure recovery capability, despite not being explicitly trained with recovery demonstrations. Whereas, for the Push Button task, it was demonstrated that proprioceptive data is essential for precise manipulation~\cite{torabi2019imitation}, acting as muscle memory and complementing the ambiguity of partial visual observations. In light of these observations, we propose a dual-branch architecture: one branch leverages visual input to automatically trigger recovery when task progress is hindered, while the other integrates proprioception with visual inputs for precise manipulation. An aggregation mechanism is designed to dynamically select and combine the outputs from both branches.

% works on the emergent network for imitation learning
\par
To mitigate the problem of shortcut learning, reinforcement learning (RL) explicitly correlates the agent's actions with the reward function, enabling it to explore the underlying causality through millions of interactions with the environment~\cite{hua2021learning, zeng2024survey}. However, the causality learned in this process would heavily depend on reward engineering. 
An inappropriate reward function can lead to poor generalization, making the learned policy less adaptable to new or unseen situations. As an alternative, Florence~\emph{et al.} proposed the implicit behavior clone (IBC), which models the policy using an energy function~\cite{florence2022implicit}. IBC learns a conditional energy-based model over actions given observations, thereby avoiding direct supervision with losses such as mean square error (MSE). In parallel, the bio-inspired neural networks (NN), which are considered as a promising direction for the next-generation neural architecture, are explored to capture the causality. Chahine~\emph{et al.}~\cite{chahine2023robust} proposed the Liquid Neural Networks (LNNs), designed to emulate the neural behaviors of the \textit{C. Elegant} animal. LLNs leverage the expressive neural ordinary differential equations (NODEs)~\cite{chen2018neural} to model the change from the input signal to the output. Despite their promise, both energy-based models and NODE-based architectures are known to be difficult to train and often struggle with scalability in high-dimensional action spaces, and incur significantly higher inference time compared to conventional neural networks. 

\par
\revision{
Recent works have explored unified modeling of video and action sequences to leverage large-scale video data in conjunction with robot actions. Li \emph{et al.}~\cite{li2025unified} introduce Unified Video Action Model (UVA), which learns a joint video–action latent representation and employs decoupled decoding with two lightweight diffusion heads, enabling real-time action inference without requiring video generation at test time. Zhu \emph{et al.}~\cite{zhu2025unified} propose the Unified World Model (UWM), which integrates video diffusion and action diffusion within a unified transformer architecture using independent diffusion timesteps, offering a scalable framework for exploiting large and heterogeneous robotic datasets. Although UVA and UWM demonstrate strong potential and promising results, these approaches rely on video prediction as auxiliary supervision, where dynamics are modeled implicitly in the image space. In this setting, the model will also account for task-irrelevant variations such as background motion, lighting, or clutter, which may introduce unnecessary complexity and might degrade overall performance.
% Unified Video Action Model~\cite{li2025unified}, Unified World Model~\cite{zhu2025unified}.
}Alternatively, to unveil the internal dynamics in demonstration datasets, latent actions have been introduced to account for structured transitions in state observations. 
% Alternatively, latent actions have been introduced to account for transitions in state observations. 
Edwards~\emph{et al.}~\cite{edwards2019imitating} proposed ILPO, which infers latent policies directly from observations by modeling the causal effect of latent actions on state transitions and then introduces an action alignment procedure to map latent actions to real-world ones. The LAPO algorithm, proposed by Schmidt~\emph{et al.}~\cite{schmidt2023learning}, models the evolution of video frames as a dynamic system, introducing pseudo-latent actions to account for temporal changes. To avoid mode collapse, these latent actions are later decoded using vector quantization. 
However, both ILPO and LAPO primarily focus on discrete action policies and require information bottlenecks to retain essential information~\cite{chen2024korol}. To explicitly incorporate latent actions into the IL framework, the data-driven tool, Deep Koopman Operator (DKO) can be adopted~\cite{lusch2018deep, shi2024koopman}. DKO provides a linear approximation of nonlinear dynamics in a latent space and has shown promise in controlling systems such as damped pendulums, CartPole, and 7-DOF robotic manipulators~\cite{shi2022deep}. Chen~\emph{et al.}~\cite{chen2024korol} extend this by using Koopman-based rollouts (KOROL) to propagate object-level features extracted from visual inputs. Unlike image-to-action policies that implicitly encode control-relevant features, KOROL explicitly constructs a Koopman operator over predicted object states to model dynamics. 
% However, it requires ground-truth object states during runtime, limiting its applicability in vision-based tasks. Additionally, 
However, KOROL computes the Koopman operator via least squares and treats the system as an autonomous model, excluding explicit actions. Thereby, Bi~\emph{et al.}~\cite{bi2024imitation} propose Koopman Operator with Action Proxy (KOAP), a plan-then-control framework that learns latent action representations from observation-only trajectories. KOAP targets continuous actions using a decision diffuser combined with Koopman-inspired inverse dynamics. It predicts future low-dimensional states and infers the latent action required to transition from the current to future states, which is then decoded into real actions. However, KOAP is computationally expensive, and its performance depends heavily on both the future state predictor and the latent action decoder. To better capture causality, in this work, we seamlessly integrate the DKO module, which models visual dynamics as a linear affine system, into the dual-branch architecture without requiring future state prediction during deployment.

\par
% action chunkings aggregation, and the feasibility of using test-time loss to weight the importance of the chunked actions
Action chunking is a pivotal design component in generative IL frameworks, such as ACT and Diffusion Policy~\cite{chi2023diffusion, zhao2023learning}, as it facilitates the modeling of temporal action dependencies within demonstrations. However, the aggregation of the overlapping ACs remains under-explored. To generate smooth and accurate command trajectories, Zhao~\emph{et al.} introduce temporal ensembling to aggregate overlapping ACs at every time step. Nakamoto ~\emph{et al.}~\cite{nakamoto2024steering} proposed selecting action samples by querying a value function learned from reward-annotated demonstrations. These policies, however, output a single AC at each inference step, with aggregation based on temporally overlapping ACs. To address scenarios where multiple ACs are produced at each inference step, Liu~\emph{et al.}~\cite{liu2024bidirectional} introduced Bidirectional Decoding (BID), a test-time decoding strategy that addresses the consistency-reactivity trade-off inherent in action chunking through sample comparison, eliminating the need for a separate value function. Although BID effectively balances this trade-off, its practical deployment is limited by factors such as increased computational cost, limited scalability concerning context length, and sensitivity to hyperparameters. To facilitate more informed aggregation, in this work, we propose to quantify the degree of out-of-distribution (OOD) of the generative AC given the environmental context.

% Pan~\emph{et al.} proposed a hybrid control method that combines the fine-tuned Vision-Language-Action model and the diffusion models. The VLA model provides language-commanded high-level planning, while the diffusion model handles the low-level interaction using a pre-trained event signal. 

\par
To balance the failure recovery capability with precise manipulation, we propose the Deep Koopman-boosted Dual-branch Diffusion Policy (D3P), a novel algorithm designed to fully unlock the potential of both visual and proprioceptive data while minimizing over-reliance on either modality. 
It features dual-branch architecture, as shown in Fig. \ref{fig:teaser}, and produces dual ACs at each inference step. 
These chunkings, along with the temporally overlapping chunkings from prior steps, are synthesized based on the uncertainties associated with the generated actions. Following the approach in~\cite{lee2024diff}, uncertainty is estimated during the test time by computing the loss function of the generative model with respect to the generated chunking. 
The main contributions of this work are summarized as follows:
\begin{enumerate}
    \item We introduce a dual-branch architecture that reduces the reliance on proprioceptive data, enabling the policy to dynamically shift focus toward visual information when appropriate. This enhances the policy’s adaptability across diverse task conditions.

    \item We integrate the Deep Koopman Operator into the proposed policy to capture visual dynamics. By preventing mode collapse through a properly designed training framework, the policy achieves enhanced visual representations and improved overall performance.

    \item To combine multiple temporally overlapping ACs from dual branches, we propose a post-processing AC aggregation module leveraging the \textit{test-time loss} of the diffusion model to signify the uncertainties of the predicted actions, which significantly improves the policy's success rate. % performance in both simulated and real environments, e.g., \todo{give numbers}.
\end{enumerate}

\par
The rest of this paper is organized as follows: Section \ref{sec:problem} formulates the problem, Section \ref{sec:method} presents the proposed method, and Section \ref{sec:exp} describes the simulation and real-world experiments, followed by an analysis of the results. Finally, Section \ref{sec:conclude} provides a discussion and conclusion.

\begin{figure*}[ht!]
\centering
\includegraphics[width=0.98\textwidth]{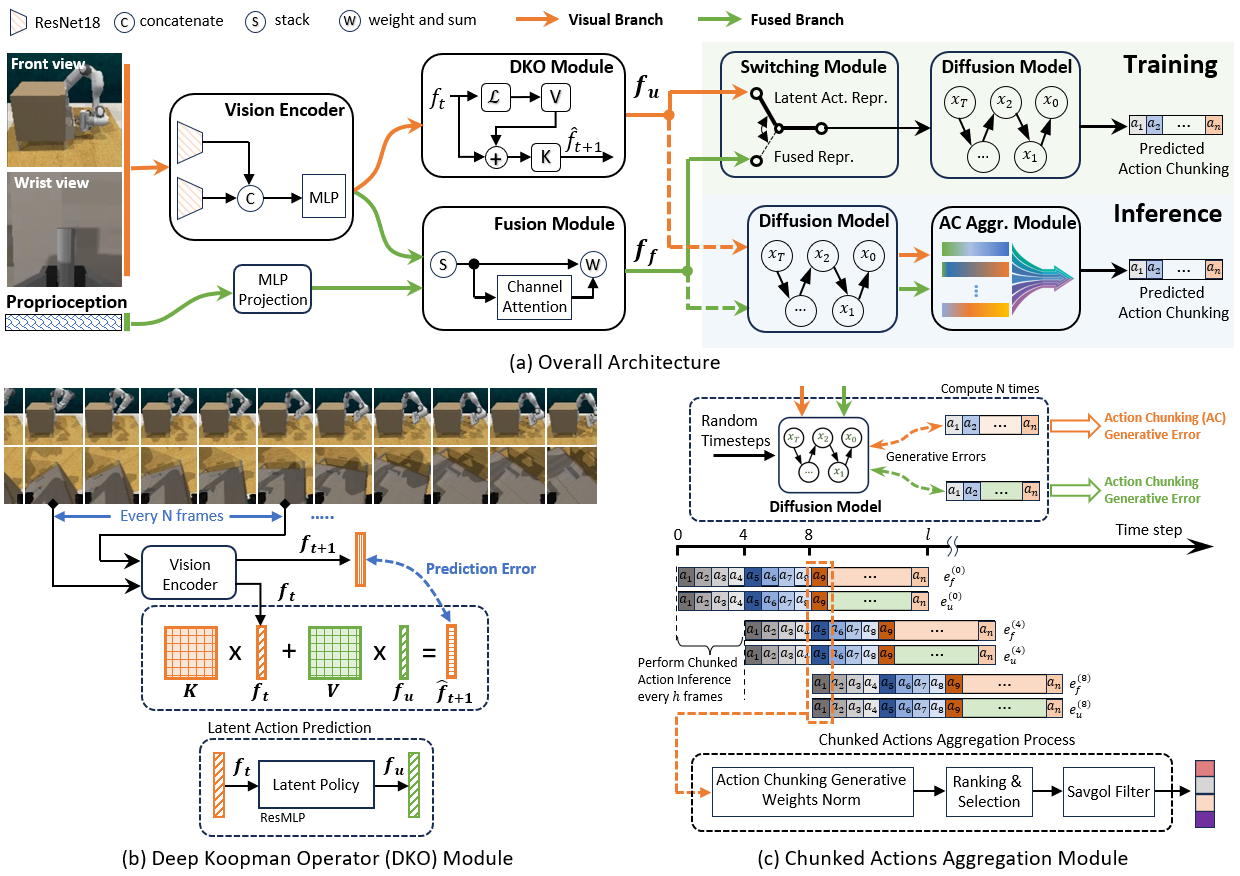}
\caption{Architecture and components of the \textit{\underline{D}eep Koopman-boosted \underline{D}ual-branch \underline{D}iffusion  \underline{P}olicy (D3P)}. (a) depicts the overall architecture of the proposed method. During training, the switching module randomly selects one of two inputs for the diffusion model: (i) $\mathbf{f}_u$, the latent action representation from (b) the DKO module, or (ii) $\mathbf{f}_f$, the fused representation of the visual and proprioceptive inputs. During inference, the diffusion model generates ACs conditioned on $\mathbf{f}_u$ and $\mathbf{f}_f$, respectively. The output ACs are then aggregated by (c) the ACs aggregation module (refer to \textit{Section~\ref{subsec:acagg}} for more details).
}
\label{fig:overview}
\end{figure*}

%%%%%%%%%%%%%%%%%%%%%%%%%%%%%%%%%%%%%%%%
\section{Problem Formulation}
\label{sec:problem}
In this section, we first define the problem within the framework of Markov Decision Processes (MDPs)~\cite{puterman2014markov}, which form the foundation of most modern robot learning algorithms, and then elaborate on the proposed method in the next section.

\subsection{Imitation Learning with Action Chunkings} 
MDPs are widely adopted to model the learning process of embodied agents. An MDP can be defined by a five-tuple $\mathcal{M}^\prime=\{\mathcal{S}, \mathcal{A}, \mathcal{P}, r, \gamma\}$, where $\mathcal{S}$ and $\mathcal{A}$ denote the state and action spaces, respectively; $\mathcal{P}(\mathbf{s}_{t+1}|\mathbf{s}_t,\, \mathbf{a}_t)$ stands for the state transition probability of the next state $\mathbf{s}_{t+1}\in\mathcal{S}$ given the current state-action pair $\left(\mathbf{s}_t,\,\mathbf{a}_t\right)$. The reward function $r:\mathcal{S}\times\mathcal{A}\rightarrow\mathbb{R}$ provides feedback to the agent upon taking action $\mathbf{a}_t=\pi(\mathbf{s}_t)$ according to a policy $\pi:\mathcal{S}\rightarrow\mathcal{A}$, while $\gamma$ is a discount factor deciding the weighting of future rewards. 

\par
Imitation learning circumvents the need for explicit reward function design in an MDP. Instead, it aims to learn an imitation policy $\pi_\Theta(\cdot)$ that replicates the expert policy $\pi_E(\cdot)$, given a dataset of expert demonstrations $\mathcal{D}=\{\tau_i\}_{i=1}^N$, where each demonstrated trajectory $\tau_i=\{\left(\mathbf{s}_j, \mathbf{a}_j\right)\}_{j=1}^{T_i}$ consists of sequential state-action pairs~\cite{gavenski2024survey}. To capture the temporal dependencies in $\mathcal{D}$, the imitation policy models the joint distribution of future actions, referred to as action chunking~\cite{lai2022action}, conditioned on past states $\pi_\Theta(\mathbf{a}_{t}, \mathbf{a}_{t+1}, \dots, \mathbf{a}_{t+l}\,|\,\mathbf{s}_{t-c}, \dots, \mathbf{s}_{t-1}, \mathbf{s}_{t})$, where $l$ and $c$ denote the length of predicted action sequence and past states at $t$-th time step as in~\cite{liu2024bidirectional}, respectively. Thus, imitation learning with action chunkings can be formulated as a four-tuple $\mathcal{M}=\{\mathcal{S}, \mathcal{A}, \mathcal{P}, \mathcal{D}\}$ and optimized using:
\begin{equation}
    \pi^* = \arg\min_{\pi_\Theta}\mathbb{E}_{(\mathbf{\bar{s}}_c,\,\mathbf{\bar{a}}_l)\sim\mathcal{D}}\Big[\,\mathcal{L}\big(\pi_\Theta\left(\mathbf{\bar{a}}_l\,|\,\mathbf{\bar{s}}_c\right), \pi_E\left(\mathbf{\bar{a}}_l\,|\,\mathbf{\bar{s}}_c\right)\big)\Big]
    \label{eq:ilopt}
\end{equation}
where $\mathbf{\bar{a}}_l^{(t)}:=\{\mathbf{a}_{t+i}\}_{i=0}^{l}$, $\mathbf{\bar{s}}_{c}^{(t)}:=\{\mathbf{s}_{t-c+i}\}_{i=0}^{c}$, and $\mathcal{L}(\cdot)$ is a loss function measuring the difference between the policies. For simplicity, we drop the superscript ``$(t)$'' in $\mathbf{\bar{a}}_l$, $\mathbf{\bar{s}}_c$ and other notations when temporal dependencies are not the focus.

\subsection{Policy Deployment with Multiple Action Chunkings}
\label{subsec:problem2}
During deployment, the learned policy executes only $h$ actions, denoted as $\mathbf{\bar{a}}_h^{(t)}$, from the predicted ACs without re-planning. The predefined horizon length $h\,(1\leq\,h\,\leq\,l)$ determines the frequency of inference, with a new prediction generated every $h$ time step. To deal with the temporally overlapping actions from past predictions, an aggregation module is required to determine the next action sequence for execution. This can be formulated as the following optimization objective:
\begin{equation}
    \mathbf{\bar{a}}_h^{(t)} = \arg\max_{\mathbf{a}_t,\dots,\mathbf{a}_{t+h}\in\mathbf{A}_h^{(t)}}\mathcal{J}\left(\left(\mathbf{A}_h^{(t)}\,,\,\mathbf{W}_h^{(t)}\right)\right)
    \label{eq:actopt}
\end{equation}
where $(\mathbf{A}_h^{(t)},\,\mathbf{W}_h^{(t)})$ is an actions-weights pair at the $t$-th time step, $\mathbf{A}_h^{(t)}$ represents a collection of temporally overlapping actions from the past and current predictions, as illustrated by the orange dashed rectangle in Fig. \ref{fig:overview} (c).
\begin{equation}
    \mathbf{A}_h^{(t)} = \mathcal{O}verlap\left(\{\dots,\,\mathbf{\bar{a}}_l^{(t-2h)},\,\mathbf{\bar{a}}_l^{(t-h)},\,\mathbf{\bar{a}}_l^{(t)}\}\right)
    \label{eq:actpool}
\end{equation}
$\mathbf{W}_h^{(t)}$, sharing the same size of $\mathbf{A}_h^{(t)}$, denotes the importance weights associated with each action in the collection. Similar to~\cite{liu2024bidirectional}, in this work, the final aggregated action sequence $\mathbf{\bar{a}}_h^{(t)}$ are selected from $\mathbf{A}_h^{(t)}$ to maximize the objective function $\mathcal{J}(\cdot)$, which is further detailed in \textit{Section \ref{subsec:acagg}}.

\section{Method}
\label{sec:method}
This section provides the details of the proposed D3P algorithm illustrated in Fig. \ref{fig:overview}. The goal of D3P is to leverage the diffusion model with both enhanced visual representation and fused representation to generate dual ACs, followed by an aggregation module to obtain a refined action sequence enhancing the robustness of the method, dealing with the potential out-of-distribution states at test time.

\par
\begin{figure*}[t!]
\centering
\includegraphics[width=0.98\textwidth]{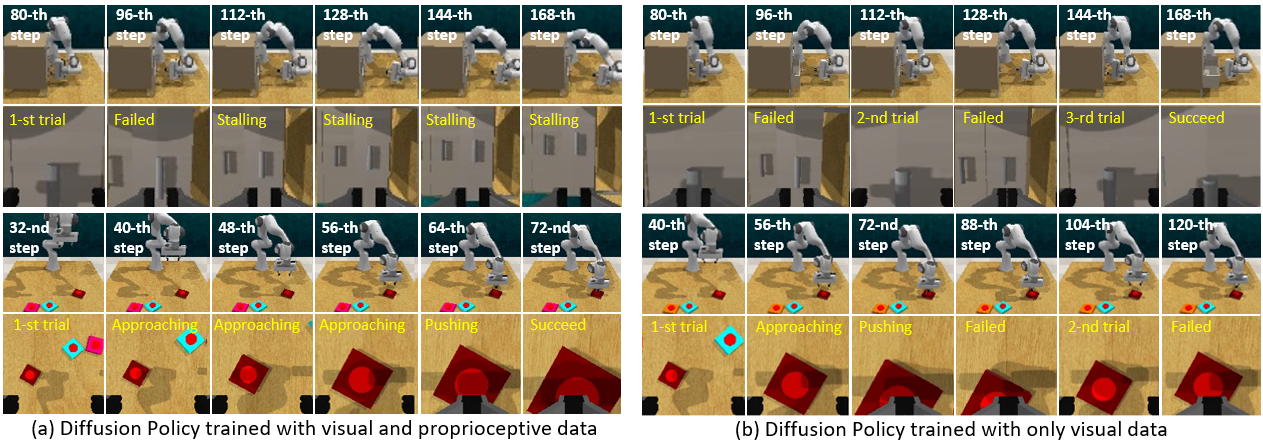}
\caption{An illustration of Diffusion Policy's performance on Open Drawer and Push Button tasks over multiple time steps when trained with (a) visual and proprioceptive data, and (b) visual data alone.
For the Open Drawer task, as shown in (a), when the robotic arm fails to grasp the drawer handle, it gradually converges to and oscillates around a fixed joint configuration, indicating a lack of recovery behavior. In contrast, in (b), under the same failure condition, the robotic arm makes repeated attempts to open the drawer, demonstrating a capacity for recovery. For the Push Button task in (b), the policy trained solely on visual data fails to determine whether the button has been pressed, due to the ambiguity of visual feedback.
}
\label{fig:illustration}
\end{figure*}

Fig. \ref{fig:illustration} intuitively presents the rationale behind leveraging dual representations. As shown in Fig. \ref{fig:illustration} (a), when the robotic arm fails to grasp the drawer handle, the policy trained with both visual and proprioceptive data encounters unseen states, leading to task failure potentially due to the shortcut learning effect~\cite{geirhos2020shortcut}. However, the policy trained with visual data alone exhibits recovery behaviors when the task progress is hindered [see Fig. \ref{fig:illustration} (b)]. This suggests that incorporating proprioceptive data can reduce the policy's robustness in certain cases. However, for the Push Button task, a policy trained with visual data alone struggles to complete the task, as it fails to distinguish whether the button has been pressed due to the inherited visual ambiguities. In such cases, proprioceptive feedback, akin to muscle memory, is crucial for precise manipulation. These observations motivate us to design an IL policy capable of generating dual ACs: one based on the fusion of all available inputs and another relying solely on visual input. These two chunkings will then be aggregated to produce the final action sequence for execution, to ensure the algorithm can better understand the situations and recover from potential out-of-distribution states. 

\par
In the following sections, we first elaborate on the encoding and training process for the fusion and visual branches, followed by the design of an aggregation module based on \textit{test-time loss} of the generative model. 

\subsection{AC Prediction based on Visual and Proprioceptive Inputs}
\label{subsec:decode}
\subsubsection{Fusing Visual and Proprioceptive Data}
\label{subsub:fusion}
The visual branch integrates the the RGB images $\mathbf{I}_t\in\mathbb{R}^{2\times3\times128\times128}$ with the proprioceptive input $\mathbf{q}_t\in\mathbb{R}^{1\times8}$, and produces a fused representation $\mathbf{f}_f\in\mathbb{R}^{1\times64}$. The RGB images are captured from the front and wrist views, while the proprioceptive input comprises the $7$-DoF joint positions of the robotic arm and the gripper width. Following the settings in ACT~\cite{zhao2023learning}, we do not incorporate past states for inference, i.e., $c=0$ in Eq. (\ref{eq:ilopt}). To process the visual input, we employ two ResNet18 networks $\mathbf{E}_\phi(\cdot)$, which are identical to those used in Diffusion Policy~\cite{chi2023diffusion}, to encode the RGB images from the front and wrist views separately. The resulting features are then concatenated and projected into $\mathbf{f}_v\in\mathbb{R}^{1\times64}$ using a multi-perceptron layer (MLP). The proprioceptive input $\mathbf{q}_t$ is first mapped to a higher dimensional space and then stacked with visual feature $\mathbf{f}_v$. Channel Attention is finally applied to balance the contributions of proprioceptive and visual representations, refining and yielding the fused representation $\mathbf{f}_f$.

\subsubsection{Action Chunking Prediction using DDPM}
The Denoising Diffusion Probabilistic Model (DDPM) ~\cite{ho2020denoising} is employed to generate ACs based on input contextual information. Here, we incorporate a switching module during training, allowing DDPM to generate ACs conditioned respectively on both the fused representation and the latent action representation $\mathbf{f}_u\in\mathbb{R}^{1\times64}$, which is introduced in \textit{Section~\ref{subsec:latent_act}}. 

\par
To learn the noise distribution, the forward diffusion process adds Gaussian noise to uncorrupted data $\mathbf{\bar{a}}_0$ progressively until it ultimately becomes white noise. Here, $\mathbf{\bar{a}}_0\in\mathbb{R}^{l\times d}$ represents an AC of length $l=16$, where the dimension of each action is $d=8$, comprising the robotic arm’s absolute joint positions and the gripper’s width. The noise data at the $k$-th diffusion step can be computed directly from the initial states by: 
\begin{equation}
    \mathbf{\bar{a}}_k = \sqrt{\bar{\alpha}_k}\,\mathbf{\bar{a}}_0+\sqrt{1-\bar{\alpha}_k}\,\epsilon
    \label{eq:dm_forwawrd}
\end{equation} 
where $\bar{\alpha}_k=\prod^{T}_{k=1}\alpha_k$, $\alpha_k=1-\beta_k$, $\epsilon\sim\mathcal{N}(\mathbf{0}, \,\mathbf{I})$, and $\mathbf{\bar{a}}_k\sim\mathcal{N}(\mathbf{\bar{a}}_k; \sqrt{\bar{\alpha}_k}\,\mathbf{\bar{a}}_0, (1-\bar{\alpha}_k)\mathbf{I})$. The maximum diffusion step is set to $K=30$, using a squared-cos noise schedule and an epsilon-type prediction. $\beta$ starts from $1\times10^{-4}$ to 0.02. For more details, one can refer to \cite{luo2022understanding, nichol2021improved}. 

The random noise distribution can thereby be learned in the reverse process by minimizing the following loss function: 
\begin{equation}
    \begin{aligned}
        \mathcal{L}_{ddpm}(\mathbf{f}_{\star},\,\mathbf{\bar{a}}_0) =& \,\mathbb{E}_{\mathbf{\bar{a}}_0, \mathbf{f}_\star,k, \epsilon}\Big[\,\big\|\epsilon-\widehat{\epsilon}\left(\mathbf{\bar{a}}_k^\star\,\big|\,k,\, \mathbf{f}_{\star},\,\theta\right)\big\|^2_2\,\Big] \\
        % \mathbf{f}_{\star} =& \,\mathbb{I}_\mathcal{S}\cdot\mathbf{f}_{a} + (1-\mathbb{I}_\mathcal{S})\cdot\mathbf{f}_{f}  \\
    \end{aligned}
    \label{eq:cddpm1}
\end{equation}
where $\mathbf{f}_{\star} = \,\mathbb{I}_\mathcal{S}\,\mathbf{f}_{u} + (1-\mathbb{I}_\mathcal{S})\,\mathbf{f}_{f}$, $k$ denotes the denoising steps. $\mathbb{I}_{\mathcal{S}}\sim Bernoulli(p)$ is a selection variable in $\{0,\, 1\}$ that follows the Bernoulli distribution and $p=0.6$. Finally, during the inference, the AC conditioned on the input contextual information $\mathbf{f}_\star$ can be denoised by iteratively running the deterministic denoising process:
\begin{equation}
    \mathbf{\bar{a}}_{k-1}^\star = \frac{1}{\sqrt{\alpha_k}}\left(\mathbf{\bar{a}}_k^\star - \frac{1-\alpha_k}{\sqrt{1-\bar{\alpha}_k}}\widehat{\epsilon}\big(\mathbf{\bar{a}}_k^\star\,\big|\, k,\,\mathbf{f}_\star,\,\theta\big)\right)
    \label{eq:denoise}
\end{equation}

\subsection{AC Prediction with Latent Action Representation}
In the visual branch, we use DKO to capture the dynamics of the visual input and enhance its representation.
\label{subsec:latent_act}
\subsubsection{Visual Dynamics Definition}
Given a primarily vision-based task, let the following equation describe the visual dynamics of each demonstrated trajectory in $\mathcal{D}$: 
\begin{equation}
    \mathbf{I}_{t+h} = \mathcal{V}(\,\mathbf{I}_{t},\; \mathbf{u}_{t})
    \label{eq-dynamic}
\end{equation}
where $\mathbf{I}_{t}$ and $\mathbf{I}_{t+h}$ denote the visual input at the current and the next time steps, respectively. The function $\mathcal{V}$, which governs the evolution of the visual stream, represents an unknown nonlinear system driven by the robot action $\mathbf{u}_{t}$. \revision{
Here, we empirically set the sampling interval $h=4$ because it can balance visual content and linearity model fidelity for the DKO module. Shorter intervals produce redundant visuals, whereas longer intervals amplify latent nonlinearity and modeling error.}

\subsubsection{Deep Koopman Operator Module}
DKO~\cite{chen2024korol} extends the Koopman framework by leveraging the deep neural network (DNN) to learn a latent space where the system dynamics become approximately linear. To account for the internal dynamics within the consecutive visual sequence, we assume that there exists an observation function $\mathbf{E}_\phi(\cdot)$, referred to as \textit{observables} in the Koopman Operator theory~\cite{brunton2019notes}, which maps the highly redundant visual data into a linear space. The transformed representation is constrained by a Koopman-like linear affine system, formulated as:
\begin{equation}
    \begin{aligned}
    \mathbf{f}^{(t+h)}_v =& \,\mathbf{K}\,\mathbf{f}^{(t)}_v + \mathbf{V}\,\mathbf{f}_u^{(t)}, \quad \mathbf{f}_u^{(t)}=\mathbf{L}_\psi\left(\,\mathbf{f}_v^{(t)}\,\right) \\
    % \mathbf{E}_\phi(\mathbf{I}_{t+h}) =& \,\mathbf{K}\,\mathbf{E}_\phi(\mathbf{I}_t) + \mathbf{V}\,\mathbf{f}_t^u \\
    \end{aligned}
    \label{eq:dko_dyn}
\end{equation}
where $\mathbf{K}$ is the Koopman Operator governing state evolution, and $\mathbf{V}$ captures the effect of control inputs. $\mathbf{f}_v^{(t)}=\mathbf{E}_\phi(\mathbf{I}_t)$ denotes the visual representation at $t$-th time step. $\mathbf{f}_u^{(t)}$ represents the latent action obtained from a causal latent policy $\mathbf{L}_\psi(\cdot)$, implemented as a ResMLP module. Notably, $\mathbf{f}_u^{(t)}$ serves as a latent action representation of the actual control input applied to the robot $\mathbf{u}_t$. This latent representation $\mathbf{f}_u$ is subsequently decoded into an AC using DDPM (see \textit{Section~\ref{subsec:decode}}). 

\subsubsection{Training Loss}
As shown in Fig. \ref{fig:overview} (b), the DKO module in this work can be considered as a linear affine model constraint imposed on the latent space of the visual dynamics. The DKO module is trained by minimizing the prediction error of the next state while avoiding mode collapse as follows:
\begin{equation}
    \mathcal{L}_{dko} = \mu\,\Big\|\widehat{\mathbf{f}}_{v1}^{(t+h)},\,\mathbf{f}_{v1}^{(t+h)}\Big\|_2^2 + (1-\mu)\,\Big\|\widehat{\mathbf{f}}_{v2}^{(t+h)},\,\mathbf{f}_{v2}^{(t+h)}\Big\|_2^2
    \label{eq:kooploss}
\end{equation}
where 
\begin{equation*}
    \begin{aligned}
        \widehat{\mathbf{f}}_{v1}^{(t+h)} &= \mathbf{K}sg\left(\mathbf{f}_{v1}^{(t)}\right) + \mathbf{V}\,\mathbf{f}_{u}^{(t)}, \; \mathbf{f}^{(\star)}_{v1} = \mathbf{E}_\phi(\mathbf{I}_{\star}), \\
        \widehat{\mathbf{f}}_{v2}^{(t+h)} &= sg\left(\mathbf{K}\right)\mathbf{f}_{v2}^{(t)} + sg\left(\mathbf{V}\,\mathbf{f}_{u}^{(t)}\right), \; \mathbf{f}^{(\star)}_{v2} = \mathbf{E}_\phi\circ\mathbf{T}_r(\mathbf{I}_{\star}), \\
    \end{aligned}
\end{equation*}
$sg(\cdot)$ represents stop gradient operation, and $\circ$ denotes function composition. $\mathbf{T}_r(\cdot)$ is an image augmentation operation applied consistently to the current and next visual input. This augmentation includes a combination of up to $\pm10\%$ cropping and padding, a $50\%$ chance of horizontal flipping, rotations up to $\pm15^\circ$, additive Gaussian noise with a standard deviation of $0.02$, and brightness adjustment within a scale of 0.9 to 1.1. The key idea is that despite variations or distortions in the images, the latent action should remain consistent, as it drives the same changes in the scene regardless of visual transformation. $\mu$ is set to $0.3$ to further enforce the output action representation to capture the transitions between consecutive observations. 

\begin{figure}[t!]
% \begin{figure}[h!]
\centering
\includegraphics[width=0.98\columnwidth]{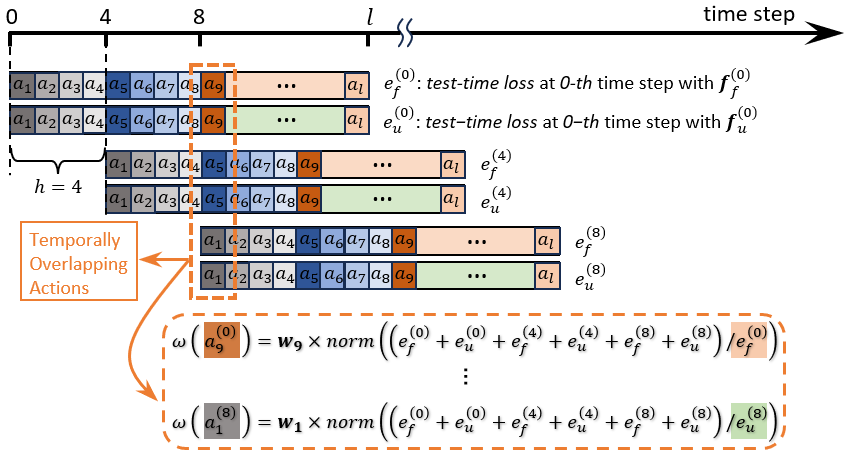}
\caption{An example for computing the associated weights for each predicted action.}
\label{fig:aggre}
\end{figure}

To summarize, the total training loss in Eq. (\ref{eq:ilopt}) for the D3P algorithm is outlined below:
\begin{equation}
    \mathcal{L} = \mathcal{L}_{ddpm} + \mathcal{L}_{dko} + \lambda\mathcal{L}_{reg}
    \label{eq:total_loss}
\end{equation}
where $\mathcal{L}_{reg}=\|\mathbf{K}\|_2^2+\|\mathbf{V}\|_2^2$ is the regularization term of the DKO module, which applies L2 weight regularization to $\mathbf{K}$ and $\mathbf{V}$. This helps improve convergence while mitigating mode collapse. Similarly, $\mathcal{L}_{ddpm}$ imposes additional constraints on the learning of $\mathbf{f}_f$ and $\mathbf{f}_u$, further preventing mode collapse in the DKO module during training. 

\begin{figure}[b!]
\centering
\includegraphics[width=0.98\columnwidth]{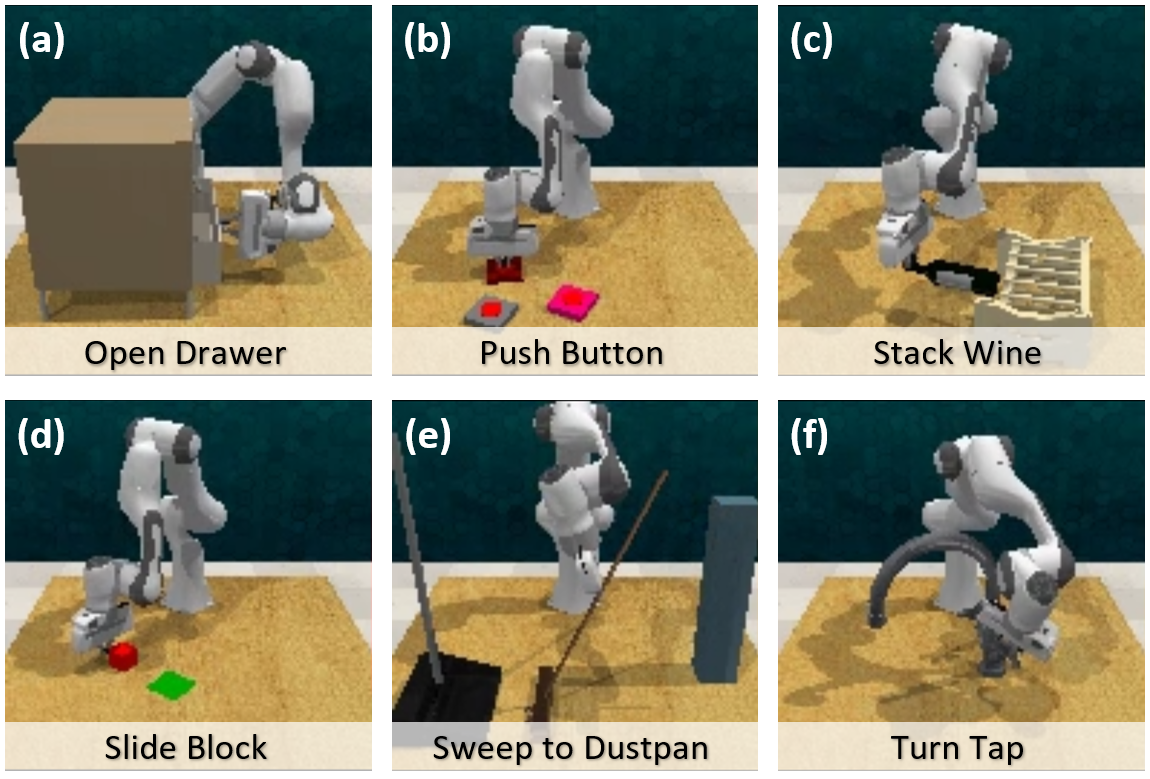}
\caption{Examples of simulated tasks selected from RLBench, including: (a) opening a drawer, (b) pushing a button, (c) stacking a wine bottle, (d) sliding a block to a target zone, (e) sweeping trash into a dustpan, and (f) turning a water tap.
}
\label{fig:simtasks}
\end{figure}

\subsection{Action Chunkings Aggregation}
\label{subsec:acagg}
To refine the final action sequence, we leverage the \textit{test-time loss} of the DDPM as in~\cite{lee2024diff} to guide the aggregation process described in Eq. (\ref{eq:actopt}). 
\begin{equation}
    e_f = \mathcal{L}_{ddpm}\Big(\mathbf{f}_f,\,\mathbf{\bar{a}}_0^f\Big),\quad e_u = \mathcal{L}_{ddpm}\Big(\mathbf{f}_u,\,\mathbf{\bar{a}}_0^u\Big) 
    \label{eq:generr}
\end{equation}
where $\mathbf{\bar{a}}_0^{f}$ and $\mathbf{\bar{a}}_0^{u}$ are the predicted outputs of DDPM conditioned on $\mathbf{f}_f$ and $\mathbf{f}_u$ using Eq. (\ref{eq:denoise}), respectively. $e_f\in\mathbb{R}^{+}$ and $e_u\in\mathbb{R}^{+}$ are referred to as \textit{test-time loss}, where higher loss indicates that the predicted action chunking deviates from the training distribution, making it less reliable. 

\par
The objective function $\mathcal{J}(\cdot)$ in Eq. (\ref{eq:actopt}) is defined: at each time step $t$, \revision{the selected action $\mathbf{a}^*_t$ maximizes the accumulation of their associated weights $\omega(\mathbf{a}_t)$:}
\begin{equation}
    \mathbf{a}^*_t = \arg\max_{\mathbf{a}_i\in\mathcal{A}_t}\omega(\mathbf{a}_i)
    \label{eq:weights_ak}
\end{equation}
where $\mathcal{A}_t$ represents the set of temporally overlapping actions. As illustrated in
Fig. \ref{fig:overview} (d) and Fig. \ref{fig:aggre}, 
the associated weight of each action is computed based on the \textit{test-time losses}, scaled by a temporal weighting factor: $\omega(\mathbf{a}_i)=w_i\times norm\left(\left(\sum_{e_i\in\mathcal{E}_t}{e_i}\right)/e_i\right)$, where $\mathcal{E}_t$ denotes the collection of \textit{test-time losses} for the overlapping ACs, and $w_i$ represents the corresponding temporal weights, given by: $w_{t+i}=\eta^i/\sum_{i=0}^{l-1}\eta^i$, $\eta=0.97$. 
\revision{During the inference, the dual-branch design can produce divergent ACs due to the jump in the action commands driving the system back to an in-distribution state. To mitigate the discontinuity and} ensure smooth execution, we apply a Savitzky-Golay filter \cite{luo2005savitzky} to refine the aggregated action sequence, effectively reducing noise while preserving the underlying motion trajectory. Pseudo-codes for the training and inference of D3P are provided in \textit{Alg. \ref{alg:training}} and \textit{Alg. \ref{alg:infer}}, respectively.

\begin{algorithm}[t]
\caption{D3P \textbf{Training}}
\label{alg:training}
\KwIn{Demo dataset $\mathcal{D}$, chunking size $l$, horizon length $h$, diffusion step $K$.}
\KwOut{Trained $\texttt{FusionModule}$, $\texttt{DKO}$, $\texttt{DDPM}$}
\ForEach{\texttt{epoch} n=1, 2, \dots}{
    \ForEach{\texttt{batch}  $(\mathbf{I}_t,\;\mathbf{I}_{t+h},\;\mathbf{q}_t,\;\mathbf{\bar{a}}_0^{(t)})\in\mathcal{D}$}{
        \mycomment{\tcc{Encode visual \& proprioceptive inputs}}
        Compute fused representation $\mathbf{f}_f \leftarrow \texttt{FusionModule}\left(\mathbf{E}_\phi(\mathbf{I}_t), \mathbf{q}_t\right)$\;
        Compute latent action representation $\mathbf{f}_u = \mathcal{L}_\psi\left(\mathbf{E}_\phi(\mathbf{I}_t)\right)$\;
        
        \mycomment{\tcc{Deep Koopman Operator prediction}}
        Compute DKO loss $\mathcal{L}_{dko}\leftarrow \texttt{Eq.(\ref{eq:kooploss})}$\;
        Compute $\mathcal{L}_{\text{reg}} \leftarrow \|\mathbf{K}\|_2^2 + \|\mathbf{V}\|_2^2$\;

        \mycomment{\tcc{Diffusion model prediction}}
        Sample diffusion step $k \sim \mathcal{U}[1, K]$\;
        Add noise to ground-truth ACs 
        $\mathbf{a}_k \sim q(\mathbf{a}_k|\mathbf{\bar{a}}_0^{(t)})\leftarrow \texttt{Eq.(\ref{eq:dm_forwawrd})}$\;
        Select representation $\mathbf{f}_{\star} = \,\mathbb{I}_\mathcal{S}\,\mathbf{f}_{u} + (1-\mathbb{I}_\mathcal{S})\,\mathbf{f}_{f}$\;
        Compute DDPM loss $\mathcal{L}_{ddpm}\leftarrow \texttt{Eq.(\ref{eq:cddpm1})}$\;

        \mycomment{\tcc{Backpropagation}}
        Update model parameters with total loss $\mathcal{L} = \mathcal{L}_{ddpm} + \mathcal{L}_{dko} + \lambda \mathcal{L}_{reg}$\;
    }
}
\end{algorithm}

\section{Experiments and Results}
\label{sec:exp}
We evaluated the proposed D3P algorithm across six simulated and three real-world tasks, comparing its performance against the state-of-the-art methods involving Diffusion Policy~\cite{chi2023diffusion} and ACT~\cite{zhao2023learning}. Additionally, we conducted an ablation study to assess the contribution of the DKO and the ACs aggregation module.

\begin{algorithm}[t]
\caption{D3P \textbf{Inference}}
\label{alg:infer}
\KwIn{
    Current observation $(\mathbf{I}_t, \mathbf{q}_t)$, horizon length $h$, Past predicted chunks $\{\mathbf{\bar{a}}_l^{(t-h)}, \mathbf{\bar{a}}_l^{(t-2h)}, \dots\}$
}
\KwOut{Executable action sequence $\mathbf{\bar{a}}_h^{(t)}$}
\mycomment{\tcc{Encode visual \& proprioceptive inputs}}
Compute fused representation $\mathbf{f}_f \leftarrow \texttt{FusionModule}\left(\mathbf{E}_\phi(\mathbf{I}_t), \mathbf{q}_t\right)$\;
Compute latent action representation $\mathbf{f}_u = \mathcal{L}_\psi\left(\mathbf{E}_\phi(\mathbf{I}_t)\right)$\;
\mycomment{\tcc{Predict action chunkings}}
Predict $\mathbf{\bar{a}}_l^f, \mathbf{\bar{a}}_l^u\leftarrow \texttt{Eq.(\ref{eq:denoise}) given } \mathbf{f}_f, \mathbf{f}_u$\;
% Predict $\mathbf{\bar{a}}_l^u\leftarrow \texttt{Eq.(\ref{eq:denoise}) conditioned on } \mathbf{f}_u$\;
\mycomment{\tcc{Aggregate action chunkings}}
Compute \textit{test-time} losses:
$e_f, e_u \leftarrow \texttt{Eq.(\ref{eq:generr})}$\;
Compute weights: \\
\ForEach{$\mathbf{a}_i \in \mathcal{A}_h\leftarrow\texttt{Eq.(\ref{eq:actpool})}$}{
    $\omega(\mathbf{a}_i) \leftarrow w_i \cdot \texttt{Normalize}\left(\left(\sum_{e_i \in \mathcal{E}_t} e_i\right) / e_i\right)$\;
}
Aggregate $\mathbf{\bar{a}}_h^{(t)} \leftarrow \texttt{Eq.(\ref{eq:weights_ak})}$\;
Apply Savitzky-Golay filter to smooth $\mathbf{\bar{a}}_h^{(t)}$\;
\Return{$\mathbf{\bar{a}}_h^{(t)}$}
\end{algorithm}

% settings
\subsection{Simulation Settings}
We employed RLBench as the testbed for the simulation study. RLBench~\cite{james2020rlbench} is a large-scale benchmark suite for robot learning that offers a diverse set of vision-based manipulation tasks within a realistic simulated environment. As shown in Fig.~\ref{fig:simtasks}, we selected six representative tasks to evaluate the proposed method. A redundant $7$-DOF robotic arm equipped with a parallel-jaw gripper and a wrist-mounted camera is employed to execute the tasks. An additional camera was positioned in the front of the table to provide a front view for the robot. The input to D3P consists of two RGB images from the front and wrist views, each with a default resolution of $128\times128$, along with the proprioceptive data comprising the robot's joint positions and gripper width at the current time step. The output of D3P is an AC of length $l=16$, containing a sequence of absolute joint positions and corresponding gripper widths. 

\par
The simulation and training were conducted on a workstation equipped with a single GPU (NVIDIA GeForce RTX 4070) and a CPU (Intel$^\circledR$ Core\texttrademark i7-13700KF Processor). For each task, we collected 80 demonstration episodes using the standard RLBench dataset generation script, with randomized task initializations and fixed initial robot joint positions. The policies were trained on these demonstrations for 200 epochs per task. After training, each model was evaluated on 40 unseen task initializations per task. To ensure a fair comparison, all models were tested on the same set of unseen task initializations.

\begin{table*}
\caption{Success Rates of Simulated Tasks under Different Methods}
\label{tab:comparison}
\centering
\tabcolsep=0.20cm
\resizebox{0.98\textwidth}{!}{
\begin{tabular}{lllllllllllllll}
\toprule
\multicolumn{1}{c}{\multirow{2}{*}{\begin{tabular}[c]{@{}c@{}}\textbf{Methods} \end{tabular}}} &
  \multicolumn{2}{c}{\textbf{\scriptsize{Open Drawer}}} &
  \multicolumn{2}{c}{\textbf{\scriptsize{Push Button}}} &
  \multicolumn{2}{c}{\textbf{\scriptsize{Stack Wine}}} &
  \multicolumn{2}{c}{\textbf{\scriptsize{Slide Block}}} &
  \multicolumn{2}{c}{\textbf{\scriptsize{Sweep to Dustpan}}} &
  \multicolumn{2}{c}{\textbf{\scriptsize{Turn Tap}}} &
  \multicolumn{2}{c}{\textbf{Average}}\\ 
\cmidrule(l){2-3} \cmidrule(l){4-5} \cmidrule(l){6-7} 
\cmidrule(l){8-9} \cmidrule(l){10-11} \cmidrule(l){12-13} 
\cmidrule(l){14-15}
\multicolumn{1}{c}{} &
  \multicolumn{1}{c}{\scriptsize{Fixed}} &
  \multicolumn{1}{c}{\scriptsize{Rand.}} &
  \multicolumn{1}{c}{\scriptsize{Fixed}} &
  \multicolumn{1}{c}{\scriptsize{Rand.}} &
  \multicolumn{1}{c}{\scriptsize{Fixed}} &
  \multicolumn{1}{c}{\scriptsize{Rand.}} &
  \multicolumn{1}{c}{\scriptsize{Fixed}} &
  \multicolumn{1}{c}{\scriptsize{Rand.}} &
  \multicolumn{1}{c}{\scriptsize{Fixed}} &
  \multicolumn{1}{c}{\scriptsize{Rand.}} &
  \multicolumn{1}{c}{\scriptsize{Fixed}} &
  \multicolumn{1}{c}{\scriptsize{Rand.}} & 
  \multicolumn{1}{c}{\scriptsize{Fixed}} &
  \multicolumn{1}{c}{\scriptsize{Rand.}} \\
\midrule
ACT~\cite{zhao2023learning}&
 ~65.0\%&  70.0\%&  60.0\%&  62.5\%&  
  62.5\%&  62.5\%&  \textbf{32.5}\%&  \underline{27.5}\%&
  35.0\%&  35.0\%&  27.5\%&  22.5\% &
  47.1\%&  46.7\% \\
DP$^*$~\cite{chi2023diffusion} &
  ~77.5\%& 40.0\%&  \textbf{97.5}\%&  \underline{77.5}\%&
  55.0\%&  22.5\%&  20.0\%&  ~5.0\%&
  \underline{70.0}\%&  15.0\%&  50.0\%&  40.0\%&
  \underline{61.7}\%&  33.3\% \\
DP$^-$~\cite{chi2023diffusion}&
 ~\underline{92.5}\%&   \underline{92.5}\%&   52.5\%&   50.0\%&
  \underline{72.5}\%&   \underline{67.5}\%&   17.5\%&   15.0\%&
  60.0\%&   \underline{52.5}\%&   \underline{55.0}\%&   \underline{57.5}\% &
  58.3\%&  \underline{55.8}\% \\
% \cmidrule(l){2-15} 
D3P (ours) &
  \textbf{100.0}\%&  \textbf{97.5}\%&  
  \underline{85.0}\%&  \textbf{80.0}\%&
  \textbf{85.0}\%&  \textbf{87.5}\%&  
  \underline{25.0}\%&  \textbf{35.0}\%&
  \textbf{87.5}\%&  \textbf{85.0}\%&  
  \textbf{75.0}\%&  \textbf{70.0}\%&
  \textbf{76.3}\%&  \textbf{75.8}\% \\
\bottomrule
\multicolumn{15}{l}{\scriptsize{The \textbf{best} and \underline{second-best} results are highlighted in bold and underlined. Fixed / Rand. denotes the robotic arm's initial \revision{joint} positions are fixed / random during evaluation.}} \\
\multicolumn{15}{l}{\scriptsize{DP$^*$ and DP$^-$ represent Diffusion Policy trained with both visual and proprioceptive data, and with visual data only, respectively.}} 
\end{tabular}
}
\end{table*}

\begin{table*}
\caption{Success Rates of Simulated Tasks with Different \revision{Action} Horizon Lengths}
\label{tab:abl_step}
\centering
\tabcolsep=0.20cm
\resizebox{0.92\textwidth}{!}{%
\begin{tabular}{lcccccccc}
\toprule
\scriptsize{\textbf{Methods}} & \scriptsize{\textbf{Horizon Length}} & \scriptsize{\textbf{Open Drawer}} & \scriptsize{\textbf{Push Button}} & \scriptsize{\textbf{Stack Wine}} & \scriptsize{\textbf{Slide Block}} & \scriptsize{\textbf{Sweep to Dustpan}} & \scriptsize{\textbf{Turn Tap}} &
\scriptsize{\textbf{Average}}\\ 
\midrule
\multirow{3}{*}{D3P visual branch \revision{only}} 
 & 16 & ~90.0\% & 42.5\% & 80.0\% & 15.0\% & 80.0\% & 47.5\% & 59.2\%\\
 & 8 & ~92.5\% & ~0.0\% & 50.0\% & 10.0\% & 80.0\% & 37.5\% & 46.7\% \\
 & 4 & ~45.0\% & ~0.0\% & 10.0\% & ~5.0\% & 67.5\% & 12.5\% & 23.3\% \\
\cmidrule(l){2-9} 
\multirow{3}{*}{D3P fused branch \revision{only}} 
 & 16 & ~85.0\% & 52.5\% & \textbf{87.5}\% & \textbf{25.0}\% & 70.0\% & \underline{67.5}\%& 64.6\% \\
 & 8 & ~75.0\% & ~0.0\% & 80.0\% & 17.5\% & 77.5\% & 55.0\%& 50.8\% \\
 & 4 & ~62.5\% & ~0.0\% & 15.0\% & 12.5\% & 55.0\% & 25.0\%& 28.3\% \\
\cmidrule(l){2-9} 
\multirow{3}{*}{D3P (ours)} 
 & 16 &  ~95.0\%& 35.0\%& 82.5\%& \underline{22.5}\%& \textbf{92.5}\%& 55.0\% & 63.8\% \\
 & 8 & ~\underline{97.5}\%& \underline{57.5}\%& 80.0\%& \underline{22.5}\%& 72.5\%& 60.0\% & \underline{65.0}\% \\
 & 4 & \textbf{100.0}\% & \textbf{85.0}\% & \underline{85.0}\% & \textbf{25.0}\% & \underline{87.5}\% & \textbf{75.0}\% & \textbf{76.3}\% \\
\bottomrule
\multicolumn{9}{l}{\scriptsize{The \textbf{best} and \underline{second-best} results are highlighted in bold and underlined.}}
\end{tabular}
}
\end{table*}

\begin{table*}
\caption{Success Rates of Simulated Tasks under Different D3P Variants}
\label{tab:ablation}
\centering
\tabcolsep=0.20cm
\resizebox{0.995\textwidth}{!}{
\begin{tabular}{lllllllllllllll}
\toprule
\multicolumn{1}{c}{\multirow{2}{*}{\begin{tabular}[c]{@{}c@{}}\textbf{Methods} \end{tabular}}} &
  \multicolumn{2}{c}{\textbf{\scriptsize{Open Drawer}}} &
  \multicolumn{2}{c}{\textbf{\scriptsize{Push Button}}} &
  \multicolumn{2}{c}{\textbf{\scriptsize{Stack Wine}}} &
  \multicolumn{2}{c}{\textbf{\scriptsize{Slide Block}}} &
  \multicolumn{2}{c}{\textbf{\scriptsize{Sweep to Dustpan}}} &
  \multicolumn{2}{c}{\textbf{\scriptsize{Turn Tap}}} &
  \multicolumn{2}{c}{\textbf{Average}}\\ 
\cmidrule(l){2-3} \cmidrule(l){4-5} \cmidrule(l){6-7} 
\cmidrule(l){8-9} \cmidrule(l){10-11} \cmidrule(l){12-13} 
\cmidrule(l){14-15}
\multicolumn{1}{c}{} &
  \multicolumn{1}{c}{\scriptsize{Fixed}} &
  \multicolumn{1}{c}{\scriptsize{Rand.}} &
  \multicolumn{1}{c}{\scriptsize{Fixed}} &
  \multicolumn{1}{c}{\scriptsize{Rand.}} &
  \multicolumn{1}{c}{\scriptsize{Fixed}} &
  \multicolumn{1}{c}{\scriptsize{Rand.}} &
  \multicolumn{1}{c}{\scriptsize{Fixed}} &
  \multicolumn{1}{c}{\scriptsize{Rand.}} &
  \multicolumn{1}{c}{\scriptsize{Fixed}} &
  \multicolumn{1}{c}{\scriptsize{Rand.}} &
  \multicolumn{1}{c}{\scriptsize{Fixed}} &
  \multicolumn{1}{c}{\scriptsize{Rand.}} & 
  \multicolumn{1}{c}{\scriptsize{Fixed}} &
  \multicolumn{1}{c}{\scriptsize{Rand.}} \\
\midrule
w/o DKO &
 ~90.0\%&  \textbf{100.0}\%&  
  \underline{87.5}\%&  75.0\%&  
  77.5\%&  70.0\%&  
  17.5\%&  \underline{32.5}\%&
  72.5\%&  80.0\%&  
  52.5\%&  \underline{65.0}\%&
  66.3\%&  70.4\% \\
\revision{D3P w/ rec. loss} &
  \revision{\textbf{100.0}\%}& ~\revision{95.0\%}&   
  \revision{\textbf{90.0}\%}&  \revision{\underline{82.5}\%}&  
  \revision{67.5\%}&  \revision{75.0\%}&  
  \revision{\textbf{30.0}\%}&  \revision{20.0\%}&
  \revision{72.5\%}&  \revision{62.5\%}&  
  \revision{52.5\%}&  \revision{47.5\%}&
  \revision{68.8\%}&  \revision{63.8\%} \\
% \revision{D3P$_{lstm}$} & 
\revision{rp. DKO w/ LSTM} &
 ~\revision{95.0\%}& ~\revision{95.0\%}&  
  \revision{85.0\%}&  \revision{\textbf{87.5}\%}&  
  \revision{55.0\%}&  \revision{60.0\%}&  
  \revision{10.0\%}&  \revision{12.5\%}&
  \revision{\textbf{95.0}\%}&  \revision{77.5\%}&  
  \revision{57.5\%}&  \revision{55.0\%}&
  \revision{66.3\%}&  \revision{64.6\%} \\
\cmidrule(l){2-15}
w/o Switch Module &
 ~87.5\%&  ~92.5\%&  
  \textbf{90.0}\%&  \textbf{87.5}\%&
  \textbf{95.0}\%&  \underline{82.5}\%&  
  \underline{27.5}\%&  22.5\%&
  \underline{92.5}\%&  \underline{90.0}\%&  
  65.0\%&  \underline{65.0}\%&
  \textbf{76.3}\%&  \underline{73.3}\% \\
rp. Aggr. Module &
 ~\underline{97.5}\%&  ~92.5\%&   
  50.0\%&  45.0\%&
  \underline{90.0}\%&  \underline{82.5}\%&   
  \underline{27.5}\%&  15.0\%&
  90.0\%&  \textbf{92.5}\%&   
  67.5\%&  \underline{65.0}\%&
  \underline{70.4}\%&  65.4\% \\
\cmidrule(l){2-15}
\revision{D3P-CM} &
 ~\revision{87.5\%}& ~\revision{90.0\%}&  
  \revision{\textbf{90.0}\%}&  \revision{\underline{82.5}\%}&
  \revision{72.5\%}&  \revision{72.5\%}&  
  \revision{25.0\%}&  \revision{17.5\%}&
  \revision{60.0\%}&  \revision{57.5\%}&  
  \revision{\underline{70.0}\%}&  \revision{57.5\%}&
  \revision{67.5\%}&  \revision{62.9\%} \\
D3P &
  \textbf{100.0}\%&  ~\underline{97.5}\%&  
  85.0\%&  80.0\%&
  85.0\%&  \textbf{87.5}\%&  
  25.0\%&  \textbf{35.0}\%&
  87.5\%&  85.0\%&  
  \textbf{75.0}\%&  \textbf{70.0}\%&
  \textbf{76.3}\%&  \textbf{75.8}\% \\
\bottomrule
\multicolumn{15}{l}{\scriptsize{The \textbf{best} and \underline{second-best} results are highlighted in bold and underlined. Fixed / Rand. denotes the robotic arm's initial \revision{joint} positions are fixed / random during evaluation.}} \\
\multicolumn{15}{l}{\scriptsize{rp. Aggr.: replace the ACs Aggregation Module with executing the AC with lower \textit{test-time loss} at each inference step, and the action horizon step is 16; }} \\
\multicolumn{15}{l}{\scriptsize{\revision{
D3P w/ rec. loss: remove the deep Koopman constraint ($\mathcal{L}_{dko}+\lambda\mathcal{L}_{reg}$) and add an image reconstruction loss.
}}} 
\end{tabular}
}
\end{table*}

\begin{figure*}[tbp]
\centering
\includegraphics[width=0.97\textwidth]{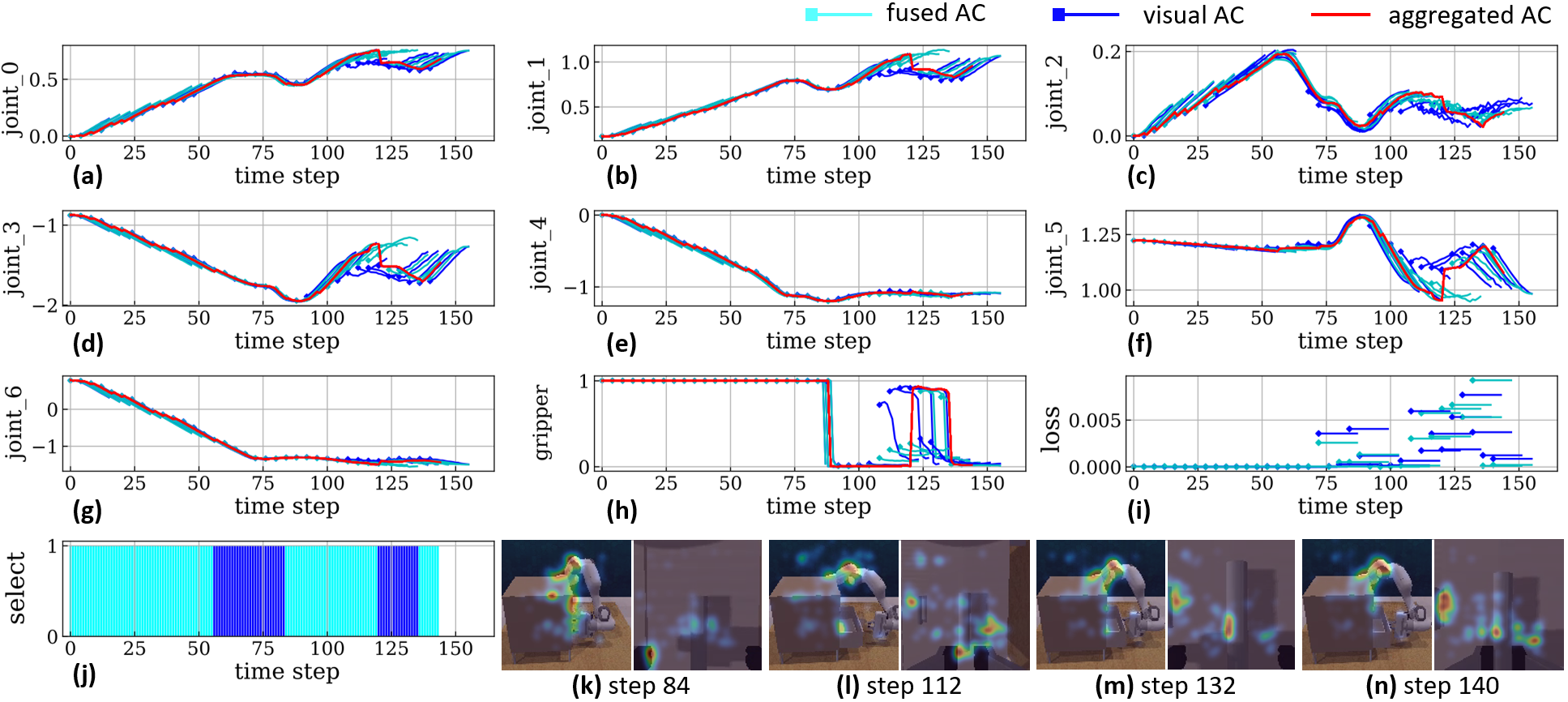}
\caption{Intermediate results for the Open Drawer task using D3P: (a)-(h) illustrate the joint position commands over time, showing the AC from both the fused and visual branches, along with their corresponding aggregated outputs. (i) plots the \textit{test-time losses} at each inference step, while (j) indicates the corresponding branch for each action, with blue and cyan bars representing the visual and fused branches, respectively. (k)–(n) display saliency maps of the input views at several stages of the task, including approaching the drawer handle, an unsuccessful grasp attempt, reapproaching, and successful completion.
}
\label{fig:intermediate}
\end{figure*}

\subsection{Simulation Results}
Fig.~\ref{fig:intermediate} shows a representative example of the intermediate results for the Open Drawer task using D3P. To better understand the model's behavior, we computed the saliency map~\cite{simonyan2013saliency} to assess the sensitivity of the encoded visual feature $\mathbf{f}_v$ with respect to the input images [see Fig.~\ref{fig:intermediate} (k)-(n)]. These saliency maps indicate that the policy primarily attends to the robot’s elbow joint and gripper as it approaches the bottom drawer handle. During the grasping phase, the focus in the wrist view shifts toward both the bottom and middle handles, suggesting their importance in estimating the success of grasping the handle required to open the drawer. During the first 70 steps, both the visual and fused branches generate similar action chunks (ACs), with consistently low \textit{test-time losses} [see Fig.~\ref{fig:intermediate} (i)]. Notable variations emerge at critical stages, such as reattempting to grasp the handle, which is influenced by visual discrepancies between the observed image and the training distribution. In the following sections, we provide a more detailed analysis of the comparisons and ablation results.

\begin{figure*}[t!]
\centering
\includegraphics[width=0.98\textwidth]{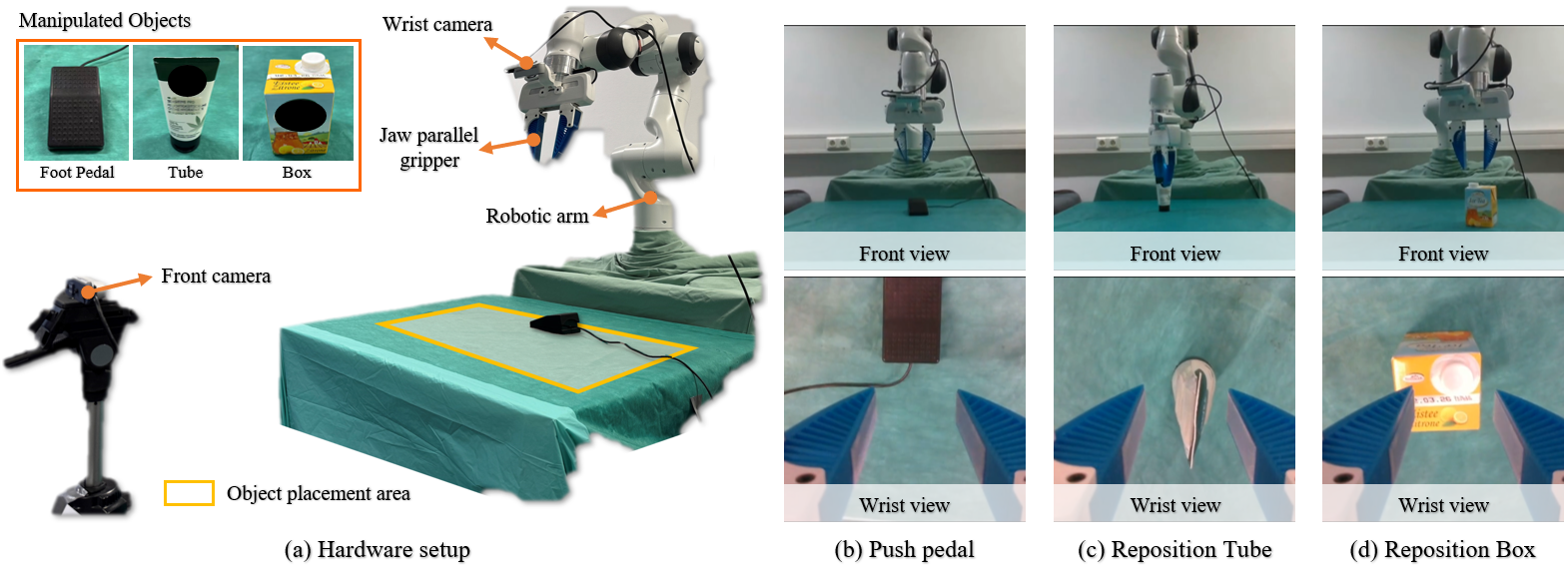}
\caption{
An illustration of the real-world experimental setup. (a). shows the task layout and initial joint configuration of the robotic arm. The three real-world tasks include: (b) pushing a foot pedal, (c) repositioning a hand cream tube, and (d) repositioning a beverage box.
}
\label{fig:realworld_tasks}
\end{figure*}

\subsubsection{\textbf{Comparison Study}}
% quantitatively  
We compared the success rates of ACT, Diffusion Policy trained with all modalities (denoted as DP$^*$), Diffusion Policy trained using only visual data (denoted as DP$^-$), and the proposed D3P, under two conditions: i). with fixed initial joint positions (``Fixed"), and ii). with randomly perturbed initial joint positions within a range of $[-10^\circ, +10^\circ]$ (``Rand."). The comparison results are summarized in Table~\ref{tab:comparison}. 
% \par
Notably, DP$^*$ shows a stronger reliance on proprioceptive data, as evidenced by a significant drop in success rate when evaluated under the ``Rand." condition (the average success rate falls from ``Fixed": 61.7\% to ``Rand.": 33.3\%). Conversely, the other methods experience less than a 3\% decrease. These results support our hypothesis that Diffusion Policy is susceptible to shortcut learning. This is further illustrated by the observation that using the same architecture but trained without proprioceptive data, DP$^-$ achieves a higher average success rate of 55.8\% under the ``Rand." condition. However, under the “Fixed” condition, DP$^*$ outperforms our proposed method in the Push Button task by a 12.5\% success rate, highlighting the importance of proprioceptive input for tasks requiring precise manipulation. Visual feedback alone appears insufficient, as analysis of the failure cases shows that under D3P control, the end-effector consistently approaches and aligns with the button’s vertical axis but fails to apply enough downward movement to push the target button. We also observe that ACT outperforms D3P in the Slide Block task (ACT: 32.5\% vs. D3P: 25.0\%). This may be attributed to the contact-rich nature of the task, which requires a larger number of demonstrations to effectively capture the environment dynamics, and a quick response to action outcomes, areas where ACT particularly excels. Despite these exceptions, our proposed D3P algorithm achieves the highest overall performance, with average success rates of 76.3\% under the “Fixed” condition (vs. 61.7\% for DP$^*$) and 75.8\% under the “Rand.” condition (vs. 55.8\% for DP$^-$), substantially outperforming the baselines.

\subsubsection{\textbf{Ablation Study}}
An ablation study was conducted to evaluate the contributions of the proposed design choices and components under the ``Fixed" condition. 
To assess the necessity of the dual-branch architecture, we executed the simulated tasks using only the ACs from either the visual branch or the fused branch. \revision{Note that in these action horizon experiments, both branches use the same horizon setting.} The results, summarized in Table~\ref{tab:abl_step}, reveal that each branch performs better on different tasks. Specifically, the visual branch outperforms the fused branch in tasks such as Open Drawer and Sweep to Dustpan. Notably, aggregating the ACs from both branches yields significantly better performance: D3P achieves an average success rate of 76.3\%, compared to 59.2\% for the visual branch and 64.6\% for the fused branch. We also observe a consistent improvement in the success rate across all single branches as the action horizon length increases from 4 to 16. For example, the visual branch improves from 23.3\% to 59.2\%. A similar trend is also observed in the fused branch. In contrast, the best performance of D3P is achieved at a horizon length of 4, which coincides with the sampling interval of the DKO in Eq. (\ref{eq:dko_dyn}) during training. Overall, these observations demonstrate the advantage of the dual-branch design.

\par
Table~\ref{tab:ablation} shows the performance of D3P when a single component is disabled. 
% On average, D3P achieves a success rate of 76.3\%, which drops notably to 66.3\% when the DKO module, \revision{shown in Fig.~\ref{fig:overview} (a), is removed and the only remaining supervision comes from the diffusion loss}. 
On average, D3P achieves a success rate of 76.3\%; \revision{when the DKO module is removed (``w/o DKO") and the model relies only on the diffusion loss,} the success rate drops to $66.3$\%. Across all six simulated tasks, D3P consistently outperforms the variant without the DKO, except for the Push Button task under the “Fixed” condition, where it performs marginally 2.5\% lower. 
\revision{We also evaluate a baseline that removes the Koopman constraint ($\mathcal{L}_{dko}+\lambda\mathcal{L}_{reg}$) and adds an image reconstruction loss (``D3P w/ rec. loss"). Compared to ``w/o DKO", it improves performance by 2.5\% under fixed initial joint configurations but degrades by 6.6\% under random initial joint configurations on average, indicating that additional visual supervision can hinder the out-of-distribution performance without structured latent dynamics.} \revision{
Furthermore, we replace the DKO module with an LSTM that can capture temporal correlations (``rp. DKO w/ LSTM'' in Table~\ref{tab:ablation}) to isolate the effect of the Koopman constraint. D3P outperforms this variant by 10.0\% on average under fixed initial joint configurations. This result indicates that DKO primarily benefits the visual branch by regularizing the visual encoder via an action-conditioned, linear-dynamics consistency constraint.
}
We also evaluated D3P without the switching module. In this setting, the DDPM loss in Eq.~(\ref{eq:total_loss}) is modified to $\mathcal{L}_{ddpm} \leftarrow 0.5\,\mathcal{L}_{ddpm}(\mathbf{f}_f, \bar{\mathbf{a}}_0) + 0.5\,\mathcal{L}_{ddpm}(\mathbf{f}_u, \bar{\mathbf{a}}_0)$. The resulting policy shows comparable or even improved performance on several tasks. Nevertheless, we still consider the switching mechanism necessary, as its exclusion can lead to reduced training efficiency. Additionally, we replaced the Aggregation Module in Fig.~\ref{fig:overview} (d) with a simpler rule: selecting the AC with the lower \textit{test-time loss} from the dual branches at each inference step and executing the full AC accordingly ($h=16$ based on the results from Table~\ref {tab:abl_step}). However, this led to a sharp performance drop in the Push Button task, probably due to the reduced adaptiveness of this rule to a changing environment. 

\par
% \todo{Analyzing the inference time breakdown and show the potential of accelerating the inference frequency.}
\revision{
To quantify inference latency, we profile the end-to-end inference time of D3P in simulation on a workstation with a NVIDIA GeForce RTX 4070 GPU and an Intel\textsuperscript{\textregistered} Core\texttrademark{} i7-13700KF CPU. As summarized in Table~\ref{tab:sim_time_inference}, diffusion denoising dominates the runtime, accounting for $248.9\pm0.9$~ms per inference, which limits D3P's applicability to dynamic or contact-rich tasks. To assess whether faster generative models mitigate this bottleneck, we distilled the diffusion model into a consistency model~\cite{song2023consistency, prasad2024consistency} while freezing the remaining trained modules, yielding the D3P Consistency Model (D3P-CM). D3P-CM uses three denoising steps and achieves an average rate of approximately 36 Hz (Table~\ref{tab:sim_time_inference}), corresponding to an approximately $9$$\times$ speedup over the approximately 4 Hz for D3P. Although D3P-CM exhibits an average performance drop of $8.8$\% relative to D3P under fixed initial joint configurations, it still outperforms DP$^{-}$ and DP$^{*}$ by $9.2$\% and $5.8$\% on average, respectively (see Table~\ref{tab:comparison} and Table~\ref{tab:ablation}).
}

\begin{table}[!t]
% \caption{\revision{Inference Time Breakdown of D3P and D3P-CM in Simulation}}
\caption{\revision{Breakdown of Inference Time for D3P and D3P-CM in Simulation Experiments}}
\label{tab:sim_time_inference}
\centering
\tabcolsep=0.20cm
\resizebox{0.485\textwidth}{!}{%
\revision{
\begin{tabular}{llrr}
\toprule
\multicolumn{1}{c}{\textbf{Parts}} &\multicolumn{1}{c}{\textbf{Components}} & \multicolumn{1}{c}{\textbf{D3P}} & \multicolumn{1}{c}{\textbf{D3P-CM}} \\
\midrule
\multirow{3}{*}{Repr. Extraction}  & Vision Encoder ($\mathbf{f}_v$) & 
3.1$\pm$0.3~ms & 3.1$\pm$0.2~ms \\
& DKO Module ($\mathbf{f}_u$) & 
0.1$\pm$0.0~ms & 0.1$\pm$0.0~ms \\
& Fusion Module ($\mathbf{f}_f$) & 
0.1$\pm$0.0~ms & 0.1$\pm$0.0~ms \\
\cmidrule{2-4}
Diffusion Model & ACs Gen. ($\mathbf{a}_0^u$, $\mathbf{a}_0^f$) & 
248.9$\pm$0.9~ms & 11.4$\pm$0.8~ms \\
\cmidrule{2-4}
\multirow{2}{*}{Aggr. Module} & Test-time Err. ($\mathbf{e}_u$, $\mathbf{e}_f$) & 
9.9$\pm$0.1~ms & 9.8$\pm$0.0~ms \\
& ACs Aggr. & 
1.2$\pm$0.0~ms & 1.3$\pm$0.0~ms \\
\midrule
% \cmidrule{2-4}
\multicolumn{2}{l}{\textbf{Average Overall Inference Time:}} & 263.2$\pm$1.2~ms & 25.7$\pm$1.6~ms \\
\bottomrule
\multicolumn{4}{l}{\scriptsize Repr.: representation. Aggr.: aggregation. Gen.: generation. Err.: errors.}\\
\multicolumn{4}{l}{\scriptsize Values are presented as: mean±std. }
\end{tabular}
}
}
\end{table}

\begin{figure*}[t!]
\centering
\includegraphics[width=0.99\textwidth]{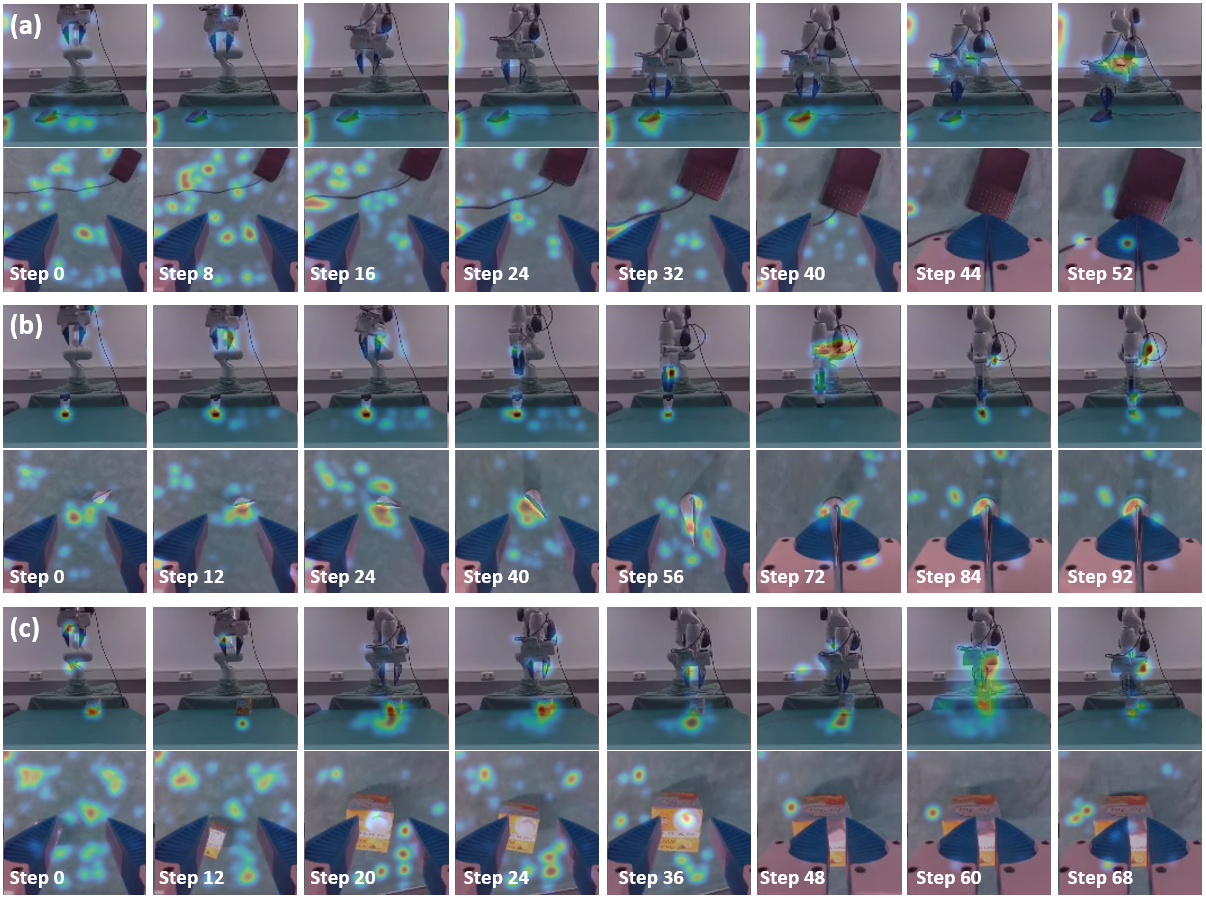}
\caption{
Saliency maps of visual inputs over time while performing three real-world tasks using D3P: (a) pushing a foot pedal, (b) repositioning a hand cream tube, and (c) repositioning a beverage box. In each subfigure, the top and bottom rows present the saliency maps for the front camera and wrist camera images, respectively.
}
\label{fig:realworld_heatmap}
\end{figure*}

\begin{table*}[t!]
% \caption{Inference Time Breakdown of D3P in Real-World Deployment}
\caption{
Breakdown of Inference Time for D3P in Real-World Experiments
}
\label{tab:time_inference}
\centering
\tabcolsep=0.25cm
\resizebox{0.985\textwidth}{!}{%
\begin{tabular}{ccccccc}
\toprule
\multicolumn{3}{c}{\textbf{\scriptsize{Representation Extraction}}} & 
\textbf{\scriptsize{DDPM}} &
\multicolumn{2}{c}{\textbf{\scriptsize{ACs Aggregation Module}}} & 
\multirow{2}{*}{\textbf{Total}}\\
\cmidrule(l){1-3} \cmidrule(l){4-4} \cmidrule(l){5-6} 
\scriptsize{\textbf{Vision Encoder ($\mathbf{f}_v$)}} & \scriptsize{\textbf{Fusion Module ($\mathbf{f}_f$)}} & \scriptsize{\textbf{DKO Module ($\mathbf{f}_u$)}} & \scriptsize{\textbf{ACs Gen. ($\mathbf{a}_0^f$, $\mathbf{a}_0^u$)}} & \scriptsize{\textbf{Test-time Err. ($\mathbf{e}_f$, $\mathbf{e}_u$)}} & \scriptsize\textbf{{ACs Aggr.}} & \\ 
\midrule
9.6$\pm$1.4~ms & 0.4$\pm$0.3~ms & 0.7$\pm$1.4~ms & 370.9$\pm$51.7~ms & 37.4$\pm$15.6~ms & 6.3$\pm$3.0~ms & 425.3$\pm$62.1~ms \\
\bottomrule
% \multicolumn{7}{l}{\scriptsize The numbers are presented as: mean$\pm$std \textbf{ms}.~~~ Calc.: calculate.~~~ Aggr.: Aggregation.}
\multicolumn{7}{l}{\scriptsize Values are presented as: mean$\pm$std. Gen.: generation. Err.: errors. Aggr.: aggregation.}
\end{tabular}
}
\end{table*}

\subsection{Real-World Settings}
% \subsubsection{Implementation Details}
In the real-world experiments, we employed a redundant robotic arm (Franka Emika Panda, Franka GmbH) equipped with a wrist-mounted camera (Intel$^\circledR$ RealSense\texttrademark D435) and a jaw parallel gripper. An additional camera (Intel$^\circledR$ RealSense\texttrademark D435) was positioned in front of the table to provide a front view of the workspace. The robotic arm was controlled via the Robot Operating System (ROS) framework, running on a laptop with an AMD Ryzen 9 5900HX CPU and an NVIDIA GeForce RTX 3070 GPU, operating on Ubuntu 20.04. As illustrated in Fig.~\ref{fig:realworld_tasks}, the robotic arm was tasked with performing three manipulation tasks: pushing a foot pedal, repositioning a beverage box, and repositioning a hand cream tube. For training, we provided 80 demonstrations per task, with the manipulated objects randomly placed on the table. Following the simulation settings, both the front and wrist images were downsampled to a resolution of 128x128. The policies were trained on these demonstrations for 200 epochs per task. To evaluate the policy performance, the objects were placed at 40 previously unseen positions for each task during testing. The placement ranges, as indicated by the yellow rectangle in Fig.~\ref{fig:realworld_tasks}, were defined as follows: Push Foot Pedal $\pm0.1~m~\times~\pm0.25~m~\times~\pm35^\circ$, and Repositioning Cream Tube and Beverage Box $\pm0.125~m~\times~\pm0.25~m~\times~\pm45^\circ$ following the format \textit{lateral range $\times$ longitudinal range $\times$ rotation range}. To reduce inference time, the number of diffusion steps in D3P was set to $K=10$. The inference time breakdown under the current setup is presented in Table~\ref{tab:time_inference}, with an average overall inference rate of approximately 2 Hz.

% data acquisition: \cite{mandlekar2022matters}.

\subsection{Real-world Results}
% \subsubsection{Results Analysis}
The success rates from the real-world experiments are presented in Table~\ref{tab:results_realworld}. A clear trend emerges with respect to task difficulty, as success rates consistently follow the order: Push Pedal $>$ Reposition Tube $>$ Reposition Box across the three evaluated methods. For instance, D3P's success rate drops from 82.5\% for the Push Pedal task to 57.5\% for Reposition Tube, and further to 25.0\% for Reposition Box. This pattern reflects the varying precision demands of each task. The beverage box repositioning task is particularly challenging, as only a narrow range of grasp poses leads to successful execution. In contrast, the hand cream tube task is more tolerant of grasping errors, resulting in higher success rates. The Push Pedal task yields the highest performance, owing to its large activation area and the uneven surface of the button, which facilitates more robust feature recognition by the policy. Notably, this contrasts with the task's comparatively lower reliability observed in simulation. In real-world settings, repositioning tasks require high-precision positioning to securely grasp objects before initiating the movement. However, the presence of noisy observations in real environments often hinders performance.

\begin{table}[t]
\caption{Results on Real-World Experiments}
\label{tab:results_realworld}
\centering
\resizebox{0.487\textwidth}{!}{%
\begin{tabular}{lcccc}
\toprule
\textbf{Methods} & \scriptsize{\textbf{Push Pedal}} &  \scriptsize{\textbf{Reposition Tube}} & 
\scriptsize{\textbf{Reposition Box}} & \textbf{Average} \\ 
\midrule
ACT~\cite{zhao2023learning} & 52.5\% & 35.0\% & 17.5\% & 35.0\% \\
% \midrule
DP$^*$~\cite{chi2023diffusion} & 55.0\% & 35.0\% & \textbf{30.0}\% & 40.0\% \\
% \midrule
D3P (ours) & \textbf{82.5}\% & \textbf{57.5}\% & 25.0\% & \textbf{55.0}\%\\
\bottomrule
\multicolumn{5}{l}{\scriptsize{DP$^*$ represents Diffusion Policy trained with both visual and proprioceptive data.}}
\end{tabular}
}
\end{table}

For the Reposition Box task, D3P exhibits a 5\% lower success rate compared to the Diffusion Policy, suggesting that the visual branch may degrade performance in tasks requiring high precision, a limitation that can be mitigated via a larger and more diverse set of demonstrations. The overall performance limitations, particularly the relatively low success rates for the Reposition Tube and Reposition Box tasks, can be attributed to several factors. These include insufficient diversity in the demonstrations, inadequate coverage of edge cases, and biased sampling. Furthermore, the reliance on only two RGB images and proprioceptive inputs restricts the model's capability to reason about the 3D spatial structure of the environment. 
The front-view saliency maps, shown in Fig.~\ref{fig:realworld_heatmap}, primarily focus on the task-relevant features, such as the target objects, while occasionally shifting attention to the robot arm's body. In contrast, the wrist-view saliency maps appear noisy during the early stage of the task but gradually converge towards the gripper as the task progresses [see wrist-view saliency maps of the first and final steps in Fig.~\ref{fig:realworld_tasks}]. Except for the Reposition Tube task, the vision encoder rarely attends to the target object in the wrist-view images. This lack of focus may contribute to the lower success rate observed in the Reposition Box task, particularly given the high precision required for this task due to the small affordance region of the beverage box. These results suggest that the vision encoder needs further improvement in spatially correlating the task-relevant features captured from both front and wrist views to enhance the overall performance.
A promising approach for future work is to explore integrating implicit 3D representations, such as multi-view transformers as used in RVT2~\cite{goyal2024rvt}, or explicit ones like voxel embeddings from PerAct2~\cite{grotz2024peract2} into the D3P framework. Changing the joint position outputs to 6-DoF Cartesian poses may also improve precision and robustness. Despite these challenges, D3P demonstrates superior overall performance, achieving an average success rate of 55.0\%, compared to 35.0\% for ACT and 40.0\% for the Diffusion Policy. This demonstrates that the proposed dual-branch AC aggregation method enhances robustness by dynamically adjusting the confidence levels of either the visual features or the fused visual-proprioceptive representations.

\section{Discussion and Conclusion}
\label{sec:conclude}
In this work, we mitigate the shortcut learning problem observed in the Diffusion Policy by introducing a dual-branch architecture designed to reduce the model's dependence on the proprioceptive data. To enhance visual representation, we integrate a DKO module, which captures the underlying visual dynamics. Additionally, an aggregation module is employed to effectively merge multiple ACs. The proposed algorithm, D3P, achieves superior performance in quasi-static tasks. However, it inherits certain limitations from the Diffusion Policy, such as slower adaptation to sudden environmental changes, which can hinder its performance in contact-rich tasks that require rapid response. Furthermore, the dual-branch structure and the use of \textit{test-time losses} for AC aggregation result in increased computational demands. Regarding latency, \revision{we realize that shorter prediction horizons increase the number of model invocations per episode and can exacerbate runtime overhead. Deep Koopman methods learn an observation map that lifts nonlinear system states into a latent space where the dynamics are well approximated by linear evolution under a learned operator. When training the DKO module, a short sampling interval yields redundant visual observations and incurs a high computational burden, whereas a long sampling interval amplifies latent nonlinearity and modeling error, making the linear approximation harder to fit and less stable. Therefore, in our setting, we empirically set a 4-step sampling interval/horizon, which offered a favorable trade-off between information content and model fidelity. With larger datasets, longer horizons should become viable, which would reduce the inference rate and thus mitigate latency, thereby shortening wall-clock task completion time.} 

\revision{To further mitigate the computation burden, we distilled the diffusion model of a trained D3P into a consistency model and evaluated it in simulation; while this substantially improved the inference rate, it incurred a measurable performance drop. More recent generative models, such as consistency trajectory models~\cite{kim2023consistency}, one-step diffusion~\cite{wang2024one}, and flow matching~\cite{yan2025maniflow}, are compatible with our framework and may further reduce inference time without compromising performance for further extension.}
\revision{
While the present study focuses on structured latent dynamics and a dual-branch recovery strategy, in the future, we will evaluate D3P on more complex scenes with more recent generative models to further validate its robustness and practical applicability.}
% \revision{While the present study focuses on structured latent dynamics and a dual-branch recovery strategy, future work can pair D3P with a faster generative model on more complex tasks to further validate its robustness and practical applicability.}
\revision{Another limitation of D3P is that action chunks are aggregated in a rule-based manner, even though we incorporate the test-time diffusion loss. This design, while simple and effective, constrains the flexibility of the aggregation process and may limit its ability to fully exploit complementary information between branches. Future work may explore hybrid approaches in which the aggregation mechanism is first parameterized and subsequently refined through limited online adaptation.}
% This issue can be alleviated through policy distillation or by adopting a one-step diffusion process to accelerate inference~\cite{prasad2024consistency, wang2024one}. \todo{admit the limitation and recall the simulation we've done, more engineering work and SOTA method for policy acceleration. cue CTM model for distillate the policy with more consistency performance without too much performance drop.}

\par
Overall, this work highlights the potential of combining complementary input modalities to overcome the over-reliance on modality-specific representations. By mitigating shortcut learning and enabling dynamic policy behavior, the proposed method lays the groundwork for more robust and generalizable imitation learning systems. While the current method is applied to a single task, future work will explore the integration of language as an additional modality, extending D3P’s capability to handle multi-task and long-horizon scenarios.

%%%%%%%%%%%%%%%%%% main content

\section*{Acknowledgments}
This work was supported in part by the Multi-Scale Medical Robotics Center, AIR@InnoHK, Hong Kong; and in part by the SINO-German Mobility Project under Grant M0221. The authors would like to acknowledge the Editors and anonymous reviewers for their time and implicit contribution to the improvement of the article's thoroughness, readability, and clarity.

\bibliographystyle{IEEEtran}
\bibliography{IEEEabrv, reference}

% \vspace{-10 mm}
\begin{IEEEbiography}[{\includegraphics[width=1in,height=1.25in,clip,keepaspectratio]{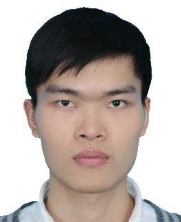}}]
{Dianye Huang} (Graduate Student Member, IEEE) received his B.Eng. degree in automation and the M.Sc. degree in control science and engineering from the School of Automation Science and Engineering, 
South China University of Technology, Guangzhou, China, in 2017 and 2020, respectively. 
% He was a junior researcher with the JiHua Lab, Foshan, China. 

He is currently pursuing his doctoral degree in computer science at the Chair for Computer-Aided Medical Procedures (CAMP) at the Technical University of Munich, Germany. His research interests include intelligent control, human-robot interaction, robot learning, medical robotics, and robotic ultrasound.
\end{IEEEbiography}

% \vspace{-10 mm}
\begin{IEEEbiography}[{\includegraphics[width=1in,height=1.25in,clip,keepaspectratio]{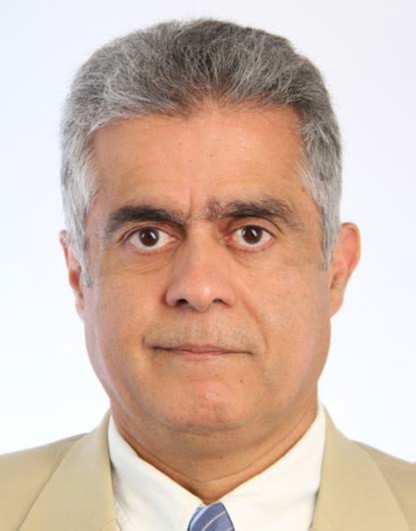}}]
{Nassir Navab} (Fellow, IEEE) received the Ph.D. degree in computer and automation from INRIA, Paris, France, and the University of Paris XI, Paris, in 1993. He is currently a Full Professor and the Director of the Laboratory for Computer-Aided Medical Procedures, Technical University of Munich, Munich, Germany, and was an Adjunct Professor at Johns Hopkins University, Baltimore, MD, USA. He has also secondary faculty appointments with the both affiliated Medical Schools. He enjoyed two years of a Post-Doctoral Fellowship with the MIT Media Laboratory, Cambridge, MA, USA, before joining Siemens Corporate Research (SCR), Princeton, NJ, USA, in 1994.

Dr. Navab is a fellow of the Academy of Europe, MICCAI, and Asia-Pacific Artificial Intelligence Association (AAIA). He was a Distinguished Member and was a recipient of the Siemens Inventor of the Year Award in 2001 at SCR, the SMIT Society Technology Award in 2010 for the introduction of Camera Augmented Mobile C-arm and Freehand SPECT technologies, and the “10 Years Lasting Impact Award” of the IEEE ISMAR in 2015.
\end{IEEEbiography}

% \vspace{-10mm}
\begin{IEEEbiography}[{\includegraphics[width=1in,height=1.25in,clip,keepaspectratio]{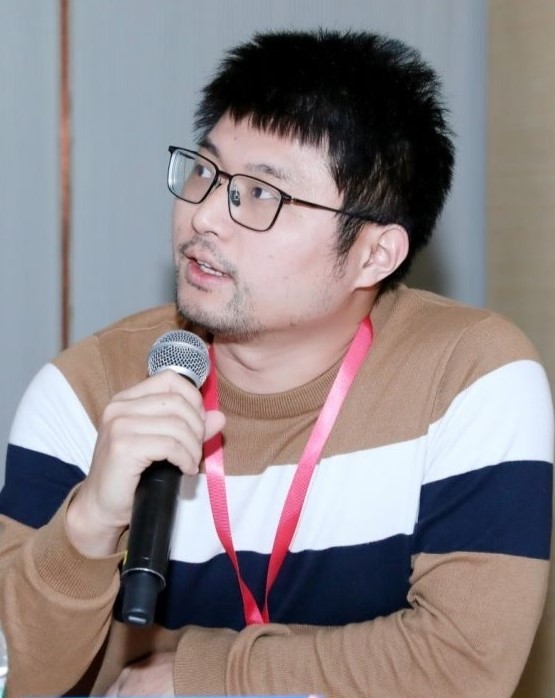}}]
{Zhongliang Jiang} (Member, IEEE) received the M.Eng. degree in Mechanical Engineering from the Harbin Institute of Technology, Shenzhen, China, in 2017, and Ph.D. degree in computer science from the Technical University of Munich, Munich, Germany, in 2022. He was a senior research scientist at the Chair for Computer Aided Medical Procedures (CAMP) at the Technical University of Munich between 2022 and 2025. From January 2017 to July 2018, he worked as a research assistant in the Shenzhen Institutes of Advanced Technology (SIAT) of the Chinese Academy of Science (CAS), Shenzhen, China. 

He is currently an Assistant Professor in the Department of Mechanical Engineering at The University of Hong Kong (HKU), where he leads the Medical Intelligence and Robotic Cognition (MIRoC) Lab. His research interests include medical robotics, robot learning, human-robot interaction, and robotic ultrasound.
\end{IEEEbiography}

\end{document}